%% file: main.tex
\documentclass{article}
\usepackage{microtype}
\usepackage{graphicx}
\usepackage{booktabs} %
\usepackage[inline,nomargin,final]{fixme}
\usepackage{enumitem}
\usepackage{times}
\usepackage{amsmath,amssymb,amsthm}
\usepackage{subcaption}
\usepackage{graphicx}
\usepackage{booktabs}
\usepackage{csquotes}
\usepackage{makecell}
\usepackage[table,x11names,dvipsnames,table]{xcolor}

\input{notation}

\usepackage{basic}
\title{Co-training Improves Prompt-based Learning for Large Language Models}

\author{Hunter Lang \\
  MIT CSAIL \\
  \texttt{hjl@mit.edu} \\\And
  Monica Agrawal \\
  MIT CSAIL\\
  \texttt{magrawal@mit.edu} \\\And
  Yoon Kim \\
  MIT CSAIL \\
  \texttt{yoonkim@mit.edu}\\\And
    David Sontag \\
  MIT CSAIL\\
  \texttt{dsontag@mit.edu}}

\begin{document}
\maketitle

\begin{abstract}
We demonstrate that co-training \citep{blum1998combining} can improve the performance of prompt-based learning by using \emph{unlabeled data}. 
While prompting has emerged as a promising paradigm for few-shot and zero-shot learning, it is often brittle and requires much larger models compared to the standard supervised setup. 
We find that co-training makes it possible to improve the original prompt model and at the same time learn a smaller, downstream task-specific model. 
In the case where we only have partial access to a prompt model (e.g., output probabilities from GPT-3 \cite{Brown2020}) we learn a calibration model over the prompt outputs. 
When we have full access to the prompt model's gradients but full finetuning remains prohibitively expensive (e.g., T0 \cite{sanh2021multitask}), we learn a set of soft prompt continuous vectors to iteratively update the prompt model. 
We find that models trained in this manner can significantly improve performance on challenging datasets where there is currently a large gap between prompt-based learning and fully-supervised models.
\end{abstract}

\section{Introduction}
\label{sec:intro}
Prompt-based learning, in which a pretrained language model is adapted to various tasks by priming on natural language prompts, has emerged as a promising framework for few-shot and zero-shot learning \cite{Brown2020,liu2021pre,wei2021fine,sanh2021multitask}. While intriguing,  these methods can be sensitive to trivial cosmetic artifacts, including variations in prompt wording and the ordering of examples \cite{lu2021fantastically,zhao2021calibrate,kumar-talukdar-2021-reordering}. Further, the models used in prompt-based learning (e.g., GPT-3, T0) are much larger than those typically used for standard fine-tuning.
These factors make prompt-based learning difficult to use and deploy in practice. 

Given a small amount of labeled data, one could evaluate the performance of each prompt and re-calibrate the prompt outputs to improve performance.
However, (i) this reliance on labeled data goes against the goal of few-shot learning, and (ii) even with oracle calibration, some prompts have sub-par accuracy.
Recently, to address issue (i), \citet{zhao2021calibrate} developed a \textit{data-free} calibration method that can dramatically improve the accuracy of few-shot prompts for GPT-3. We build on their work by showing how to use \textit{unlabeled data} to further improve performance.

To leverage unlabeled data, we use \textit{co-training} \citep{blum1998combining}, which operates on two views of each data point $X$: $\phi_0(X)$ and $\phi_1(X)$.
For example, in a clinical diagnosis system, $\phi_0(X)$ could be laboratory test results and $\phi_1(X)$ an X-ray image. 
A pair of models ($h_0$ and $h_1$ respectively) takes turns labeling a large unlabeled training set, and each model is trained on the confident pseudo-labels from the other.
Model $h_0$ only uses $\phi_0(X)$, and model $h_1$ uses $\phi_1(X)$.
By using complementary information in the views $\phi_0, \phi_1$ and the different inductive biases from models $h_0$, $h_1$, co-training allows each model to learn from the other without the need for labeled data.
The initial signal to start the co-training process is provided by a ``guess'' at a model $h_0$.
To combine co-training and prompt-based learning, we use outputs from a large prompt-based model as $\phi_0(X)$ and the pre-trained representation from a much smaller language model (e.g., DeBERTa \citep{he2021deberta}) as $\phi_1(X)$. We specify the models $h_0$ and $h_1$ based on whether we have partial access to the prompt model (querying GPT-3) or full access (locally training T0).

\begin{figure*}[t]
     \centering
         \includegraphics[width=\textwidth]{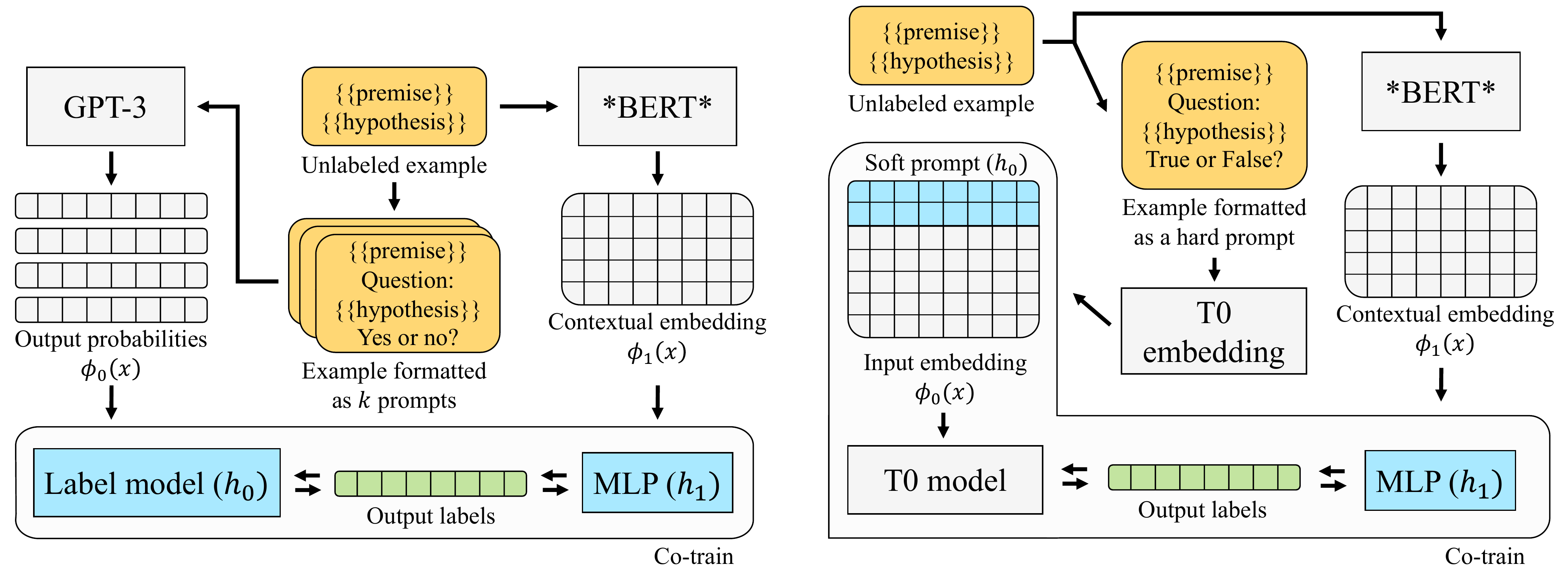}
         \label{fig:gpt3-cotrain}
    \caption{The setup for our two applications of co-training to prompting for a binary entailment classification dataset (RTE). Parameters in blue are trainable; models in gray are fixed. Left: training a ``label model'' for post-hoc calibration and ensembling of multiple prompts. Here the prompts and the model (GPT-3) are fixed, and we co-train the calibration / ensembling parameters with the task-specific model (e.g., DeBERTa). 
    Right: training a soft prompt. Here the input is encoded as a hard prompt and the embedding matrix of the input sequence is obtained. A $L \times d$ matrix of trainable parameters (the ``soft prompt'') is prepended to this embedding, and the combined embedding sequence is passed through T0 to get output predictions. We co-train the soft prompt with the view 1 model (e.g., DeBERTa).}
    \label{fig:cotraining}
\end{figure*}

In \emph{partial access}, we only have access to the large model's output probabilities.
In this case, we use unlabeled data to learn a model $h_0$ that both calibrates individual prompts and ensembles multiple prompts. We refer to this as the \emph{label model}.
We use Calibrate-Before-Use \cite{zhao2021calibrate} to initialize the calibration parameters of this model for each prompt, and we initialize the ensembling parameters to approximate majority vote.
We then refine this initial guess for $h_0$ with co-training.
We use the pre-trained representation from DeBERTa \citep{he2021deberta} for $\phi_1(X)$ and train the last few layers of that model as $h_1$.
The only labeled data used is the set of $k$ examples used in the input prompts.
Figure~\ref{fig:cotraining} (left) shows the co-training process in the partial access setting.

We also study a \emph{full access} setting using T0 \citep{sanh2021multitask} instead of GPT-3, so we can introspect the large prompt model.
We derive the view $\phi_0(X)$ and the model $h_0$ from T0.
However, instead of fully fine-tuning T0 during co-training, we focus on \emph{soft prompt tuning}, which trains several orders-of-magnitude fewer parameters while attaining similar performance \citep{li-liang-2021-prefix, lester-etal-2021-power}.
The parameter space for model $h_0$ is the set of \emph{soft prompts}, which are matrices $\mathbb{R}^{L \times d}$, where $L$ is a sequence length hyperparameter and is $d$ the dimension of the pretrained T0 embeddings.
Each row of the soft prompt mimics the embedding of a token, but the soft prompt need not correspond to the embedding of any actual token sequence.
This matrix is prepended to the input embedding and the output of $h_0$ is computed with the frozen T0 model.
The initial guess at $h_0$ (i.e., the initial soft prompt vector for use in co-training) is the repeated embedding of the \texttt{[PAD]} token.
Since T0 was trained to perform well at zero-shot learning with prompts, this provides a good initial hypothesis.
We co-train this model with a pre-trained DeBERTa representation as $\phi_1(X)$ and the last few layers of DeBERTa as $h_1$.
This is is shown in Figure \ref{fig:cotraining}, right.

We apply our approach to standard few-shot and zero-shot benchmarks and find that (i) iteratively co-training both models using unlabeled data consistently improves performance, (ii) pseudo-labels from a prompted model are an effective signal for fine-tuning smaller task-specific models, and (iii) this approach can significantly improve results on datasets previously considered difficult for prompt-based learning. 
We conclude with a brief analysis of success and failure cases and describe high-level criteria required for our method to work.

\section{Related work}
\paragraph{Prompting and prompt tuning.}

\citet{lu2021fantastically} find optimal orderings of prompt examples based on an artificially constructed development set.
Given the variance in performance across different prompts, others have focused on engineering suitable prompts, manually or otherwise \citep{liu2021pre}. \citet{jiang2020can}, \citet{shin-etal-2020-autoprompt}, and \citet{gao2020making} use data-driven techniques and language models to automatically generate candidate prompts. Rather than being constrained to human-readable prompts, \citet{li-liang-2021-prefix} and \citet{lester-etal-2021-power} instead learn a continuous \textit{soft} task-specific ``prompt" to condition language models. While effective, these methods typically require nontrivial amounts of labeled data.

Another line of work uses the outputs from a prompted language model as weak labels, as we do in this work. \citet{wang-etal-2021-want-reduce} propose to train smaller models on labels from GPT-3 to reduce annotation cost, but they train from individual, uncalibrated prompts and do not attempt to refine the prompt model alongside the smaller model.
\citet{schick2021exploiting} fine-tune a separate RoBERTa model for \emph{each} prompt using a small amount of labeled data.
They next aggregate the outputs of these individual fine-tuned models as a soft pseudo-label and train a final model to match the soft aggregation.
In contrast, we train a single BERT-style model on the ensembled prompt output without any additional labeled data.
We use this model to refine the ensemble parameters (and vice-versa).
In our approach we only use prompt outputs as training signal, and we consider different types of prompts (open-ended instead of cloze).

\paragraph{Self-training for few-shot text classification.}
Our work relies on access to a large amount of unlabeled data to iteratively grow a confidently-labeled training set for each model. 
Similarly, \textit{self-training} first trains a model on a small set of initial data, uses the trained model to produce pseudo-labels on a set of unlabeled data, and then iteratively includes the confidently pseudo-labeled data as new training labels \citep{scudder1965probability}.
In the context of few-shot text classification, \citet{mukherjee2020uncertainty} develop an uncertainty-aware technique for choosing which data points to include, which requires a small amount of labeled data.  \citet{karamanolakis-etal-2019-leveraging, karamanolakis2021self} 
  employ self-training and iterative co-training with weak supervision as the initial label signal, and they similarly use a neural network with pretrained embeddings as a downstream model. However, they explore hand-written or keyword-based rules as weak supervision, in contrast to the present work, where we derive our weak signals from prompted models. 
  The parameterization of $h_0$ in our partial access setting is similar to the weighting they use to combine rules.

\paragraph{Co-training.}
Co-training dates back to \citet{blum1998combining}, who assumed that $\phi_0(X)$ and $\phi_1(X)$ are two distinct views and conditionally independent given the true label.
Under this strict condition, they proved that the algorithm finds a good classifier after just one step.
Many subsequent analyses \citep[e.g.,][]{dasgupta2002pac, balcan2005co} relax this condition, showing that views can be dependent or even identical as long as certain relationships hold between the models being trained (essentially, they are ``different enough'').
In a similar vein, \citet{wei2020theoretical} give a theoretical explanation of why (and when) models can learn to be more accurate than the pseudo-labels used to train them.
We take implicit advantage of these results in our work.
The views we use are highly dependent, and yet the models we train are often able to outperform the pseudo-labels we used to train them in each co-training iteration.

\section{Co-training with prompting}
\label{sec:co-training-desc}
The skeleton of our approach is shown in Algorithm \ref{alg:cotrain} (full detail is provided in Algorithms \ref{alg:cotrain-gpt3-detailed} and \ref{alg:cotrain-t0-detailed} in the supplement).
First, a hypothesis $h_0$ over view $\phi_0$ is initialized such that its initial predictions are reasonable. (We discuss initialization in depth in the following sections.)
Next, we obtain the confidently labeled training data $L_0^0$, which is a subset of the unlabeled data points, together with pseudo-labels for those points from $h_0$.
In iteration $t$, we select a $\beta + t\beta'$ fraction of the data. (We discuss techniques for selection of confident data in Section~\ref{sec:confident} and the choice of $\beta$ and $\beta'$ in Section~\ref{sec:experiments}.)
These confidently-labeled points are then used to train a model $h_1$ on view $\phi_1$, and $h_1$'s confidently-labeled data is extracted as $L_1^0$.
This is used to train a new $h_0$, and the process continues for $T$ steps.
\texttt{Train} performs standard supervised training on the pseudo-labels for that iteration.
In this section, we give details for how to construct the views $\phi_0$ and $\phi_1$, the hypothesis classes we use for the model $h_0$, and the initialization schemes for $h_0$ in both the partial access and full access settings.

\begin{algorithm}[t]
  \caption{Co-training algorithm}
  \label{alg:cotrain}
  \begin{algorithmic}
  \INPUT $\mathcal{U} = \{x_n\}_{n=1}^U$ unlabeled examples
  \INPUT $\{(x_j, y_j)\}_{j=1}^k$ labeled examples (optional)
  \INPUT initial coverage $\beta$, coverage increase $\beta'$  
  \STATE $h_0 \gets$ \texttt{InitClassifier}($\phi_0$)
  \FOR{$t \mbox{ in } \{0,\ldots, T-1\}$}
      \STATE $\tilde\beta \gets \beta + t\beta'$
      \STATE {\color{gray} // \texttt{GetConfData}$^*$ defined in Algorithms \ref{alg:get-conf-data-mc}, \ref{alg:get-conf-data-cs}}
      \STATE $L_0^t \gets$ \texttt{GetConfData}$^*\left(\mathcal{U};\ h_0, \phi_0, \tilde\beta\right)$
      \STATE $h_{1} \gets$ \texttt{Train}($\phi_{1}, L_0^t$)
      \STATE $L^t_1 \gets$ \texttt{GetConfData}$^*\left(\mathcal{U};\ h_1, \phi_1, \tilde\beta\right)$
      \STATE $h_{0} \gets$ \texttt{Train}($\phi_{0}, L_1^t$)
  \ENDFOR
  \STATE {\bfseries return} $(h_0, h_1)$
\end{algorithmic}
\end{algorithm}

\subsection{Partial access: co-training a label model}

In the usual few-shot setting with prompting, $k$ labeled examples $(\{x_i, y_i\})_{i=1}^k$ are converted into a single natural language prompt following a prompt template.
We call this prompt $k$-shot, since it uses $k$ labeled examples.
Instead of using one $k$-shot prompt, we use $k$ one-shot prompts, only including one example in the template at a time. This gives us $k$ outputs.
Separating out the signal from each labeled example in this way allows us to combine the examples more effectively than the one $k$-shot prompt model. 

\paragraph{View.} Let $\phi^{(i)}_0(x) \in \mathbb{R}^{|V|}$ be the vector of probabilities output by GPT-3 on input $x$ formatted in a one-shot prompt with labeled example $(x_i, y_i)$.
Here $V$ is a subset of the full token vocabulary---the \emph{verbalizer tokens}---and consists of the ``label tokens'' for the prompt as well as other tokens related to the label.
For example, in sentiment analysis, if $x_1$ is ``this movie was great!'', $\phi^{(1)}_0(x)$ is GPT-3's output on:

\begin{displayquote}
\small
\texttt{Review: this movie was great!}\\
\texttt{Positive or Negative? Positive}\\
\texttt{Review: \{\{$x$.review\}\}}\\
\texttt{Positive or Negative?} 
\end{displayquote}
\vspace{1mm}
and $V$ might include the label tokens \texttt{\small Positive} / \texttt{\small Negative} and related tokens such as uncased label tokens or synonyms.\footnote{In the running sentiment analysis example we might have $V = \{${\scriptsize \texttt{Negative}, \texttt{Positive}, \texttt{negative}, \texttt{positive}, \texttt{bad}, \texttt{good}, $\ldots$}$\}$.}  To select the verbalizer tokens in a task-agnostic way, we obtain the top 10 predictions for GPT-3 on each prompt, sort tokens by the total probability assigned to them on the unlabeled training set, and choose the top 25\%. This ensures that other frequent tokens appear in the feature set for $\phi_0$. For example, \texttt{\small Date} appears for TREC question classification even though the closest true label category is \texttt{\small Number}. The label model automatically learns this association during the co-training iterations. 
Henceforth we assume that $V$ is ordered and the first elements of $V$ correspond to the label tokens.
By concatenating $\phi_0^{(i)}(x)$ for each of the $k$ labeled examples, we obtain a matrix $\phi_0(x) \in \mathbb{R}^{k\times |V|}$, the first view for co-training.
The second view, $\phi_1(x)$, is the frozen representation of a pretrained model like DeBERTa \cite{he2021deberta}.
In our experiments, we use the representation in the penultimate layer as $\phi_1(x)$, and the hypothesis class over this view is the last layer and the linear classifier. 

\paragraph{Hypothesis class.} This leaves the hypothesis class for model $h_0$: how do we combine $k$ prompt signals into one pseudo-label? 
Probabilities from these models are often miscalibrated \cite{zhao2021calibrate}, and thus averaging or majority voting does not yield good results.
Instead, we propose to learn a \emph{label model} that scales each prompt vector $\phi_0^{(i)}(x)$ by a prompt-specific \emph{calibration matrix} $W^{(i)}$ before averaging.
The combined architecture for this model is given by $h_0(x; W,\alpha)$:\footnote{Here we use $W$ to refer to $\{W^{(i)}\}_{i=1}^k$.}
\begin{equation}
    \label{eqn:linear-label-model}
    \begin{split}
        {{\bf l}}_i &= \ReLU\left(W^{(i)} \phi^{(i)}_0(x)\right);\\ 
        h_0(x; W, \alpha) &= \softmax\left(\sum_{i=1}^k \alpha_i {{\bf l}}_i\right),
    \end{split}
\end{equation}
where $\alpha \in \mathbb{R}^k$ is a vector of weights for ensembling the scaled prompt outputs.
The $\ReLU(\cdot)$ allows the model to easily ignore particular prompt/label combinations.
For example, if prompt $j$ has very poor precision when it outputs label $z$, setting $W^{(j)}_{zz}$ to be negative causes $l_{jz}$ to be~0.
Note that we directly calibrate probabilities rather than \emph{log}-probabilities, following \citet{zhao2021calibrate} (i.e., $\phi_0^{(i)} \in [0,1]^{|V|}$ and $\Vert \phi_{0}^{(i)} \Vert_1 = 1$).
In each iteration of co-training, we train this model using the standard cross-entropy loss on the confident data for that iteration.

\paragraph{Initialization.} %
The calibration matrices $W^{(i)}$ are initialized using Calibrate-Before-Use (CBU) \citep{zhao2021calibrate}, which first computes the probability vector $\phi_{0}^{(i)}(x_{cf})$ on  content-free inputs $x_{cf}$ (e.g., \texttt{N/A} or the empty string), and then uses these as a scaling factor:
\[
W^{(i)} = \diag\left(\frac{\mathbf{1}}{\phi^{(i)}_{0}(x_{cf})}\right).
\]
This initialization scheme ensures that the scaled prompt outputs are neutral on truly neutral inputs, improving calibration.
We also set $\alpha_i = 1$ for each $i \in \{1,\ldots, k\}$ to initially weight each prompt equally in the ensemble.
When $V$ consists of more than just label tokens, we initialize $W^{(i)}$ in blocks: we use CBU for the label tokens, and initialize the rest of $W$ to be 0.
That is, for an $l$-way classification task, where $W^{(i)} \in \mathbb{R}^{l \times |V|}$, we initialize the first $l$ columns of $W^{(i)}$ using CBU (since we assumed $V$ was ordered, these columns correspond to the label tokens) and set the rest to 0.
Hence only the label tokens (e.g., \texttt{\small Negative}, \texttt{\small Positive}) are used at initialization, but subsequent iterations of co-training can use nonzero weights on the extra tokens.
\subsection{Full access: co-training a soft prompt}
In this setting, our prompt model is the T0 model \citep{sanh2021multitask}, which achieves zero-shot generalization by fine-tuning T5 \cite{raffel2020exploring} on multiple tasks whose labeled examples have been transformed into natural language question-answer pairs.
Since T0 is publicly available and smaller than GPT-3, we can introspect the model and compute gradients in this case. 

\paragraph{View.} We set $\phi_0(X)$ to the initial word embeddings of T0 and leave $\phi_1(X)$ and $h_1$ unchanged (i.e., $\phi_1$ is the penultimate layer of a pretrained DeBERTa representation). %

\paragraph{Hypothesis class.} The model $h_0$ is parameterized by a continuous \emph{soft prompt} \cite{li-liang-2021-prefix, lester-etal-2021-power}.
Concretely, letting $d=2048$ be the dimension of the T0 word embeddings,
a soft prompt is a matrix of parameters $P \in \mathbb{R}^{L\times d}$, where $L$ is a sequence length hyperparameter.
Each row of the soft prompt acts like the embedding of a ``token'' (but needn't correspond to the embedding of any real token---i.e., there are no constraints on $P$). The hypothesis $h_0(x; P)$ is thus given by prepending the soft prompt to the input word embedding sequence and using the concatenation $(P; \phi_0(X))$ as input to $T0$.
The subsequent T0 layers are frozen and not updated during training.
Given enough labeled data, soft prompt tuning can match the performance of full-fine-tuning with far fewer trainable parameters \citep{lester-etal-2021-power, le-scao-rush-2021-many}.\footnote{We use $L=20$ in our experiments, following \citet{lester-etal-2021-power}, so the soft prompt has $20 \times 2048 = 40960$ parameters.}

\paragraph{Initialization.} T0 is specifically trained to perform well at zero-shot tasks with a variety of hard prompts, so using a hard prompt out-of-the-box gives good initial performance. 
Hence, to initialize a \emph{soft} prompt hypothesis, we encode the input using a hard prompt and then set the soft prompt to be the repeated embedding of the tokenizer's padding token. Using the RTE dataset \citep{dagan2005pascal} as a running example, we first encode the input using a hard prompt, where each input example $x$ is formatted as:
\vspace{1mm}
\begin{displayquote}
\small
\texttt{\{\{$x$.premise\}\}\\}
\texttt{Question: \{\{$x$.hypothesis\}\} True or False?}
\end{displayquote}
\vspace{1mm}
We then set $h_0$ to be the repeated embedding of the T0 padding token, i.e., at initialization the T0 model sees:
\vspace{1mm}
\begin{displayquote}
\small
\texttt{[PAD]...[PAD]\{\{$x$.premise\}\}\\}
\texttt{Question: \{\{$x$.hypothesis\}\} True or False?}
\end{displayquote}
\vspace{1mm}
This combination of hard prompt encoding with soft prompting differs from the usual soft prompting setup \citep{li-liang-2021-prefix, lester-etal-2021-power}. 
We discuss this issue in more depth in Section \ref{apdx:sp-encoding}.

\section{Selecting confident data}
\label{sec:confident}

The key step in co-training is selecting confidently-labeled data for use in the next training iteration.
The literature on co-training has identified a large number of methods for performing this data selection (\texttt{GetConfData}, in Algorithm \ref{alg:cotrain}).
We consider two simple approaches in this work: \emph{model confidence} and \emph{cut statistic}.
In both cases, we specify an initial coverage fraction $\beta$ and a coverage increase fraction $\beta'$.
Given $U$ unlabeled examples, the amount of pseudo-labeled data in round $t \ge 0$ is therefore $U(\beta + t\beta')$.

\paragraph{Model confidence.}
For model confidence, we sort every example by the scores output by each model and select the top $\beta + t\beta'$ fraction in iteration $t$.
While simple, this can result in very imbalanced updates to the pseudo-labeled dataset if the model is only confident for one label or if one label is inherently more noisy than the others.
If additional knowledge regarding the marginal label distribution is available (e.g., approximate label balance or a constraint on the minimum label frequency), we can imbue this knowledge into the data selection process by grouping examples by their predicted label and then performing the sort-and-select procedure for each label separately.
Knowledge of the approximate label balance is a standard assumption in weak supervision \citep[e.g.,][]{fu2020fast}, but we make a much weaker assumption when using the model confidence ranking: we assume we know a \emph{lower bound} $\gamma$ such that for all labels $y$, $\P[Y=y] \ge \gamma$. 
We set $\gamma=0.01$, i.e., that every class accounts for at least 1\% of the data.
The detailed procedure for confident data selection using model confidence is shown in Algorithm \ref{alg:get-conf-data-mc} (supplement).

\paragraph{Cut statistic.}
The cut statistic is a ranking heuristic that uses the view geometry more than the model confidence approach \cite{muhlenbach2004identifying, zhang2011cotrade}.
Suppose we want to select data confidently labeled by a model over view $\phi(X)$ (we omit the subscript $i$ for clearer notation).
First, we form a graph $G = (V,E)$ with one vertex for each unlabeled training example and edges connecting vertices who are $K$-nearest neighbors in $\phi(X)$ (or a representation related to $\phi(X)$---for example, for T0 we can use a contextual representation from inside the model instead of the uncontextual embeddings $\phi_0$).

Let $\hat{Y}(X) = \arg\hspace{-0.3mm}\max \, h(\phi(X))$ be the hard pseudo-label assigned to input $X$ by model $h$.
We say an edge $(x_u, x_v)$ is \emph{cut} if $\hat{Y}(x_u) \ne \hat{Y}(x_v)$.
Intuitively, we can feel confident about examples that have few cut edges, since they have the same label as most of their neighbors.
Regions of $G$ with high noise are less likely to be correctly labeled.
The cut statistic heuristically quantifies this idea to rank examples.

Suppose (as a null hypothesis) that the labels $\hat{Y}$ were sampled i.i.d. from the marginal distribution $\P[\hat{Y} = y]$ (i.e., independently of $X$).
For vertices $u$ and $v$ corresponding to examples $x_u$, $x_v$, define 
$I_{uv} = \mathbb{I}[\hat{Y}(x_u) \ne \hat{Y}(x_v)].$
Consider the test statistic:
$
J_u = \sum_{v\in N(u)}w_{uv}I_{uv},
$
where $w_{uv} = 1/(1+\Vert\phi(x_u) - \phi(x_v)\Vert_2)$ are edge weights that decrease as the distance between $u$ and $v$ increases, and $N(u)$ are the neighbors of $u$.
The mean of $J_u$ under the null hypothesis is: 
$\mu = (1-\P[\hat{Y}(x_u)])\sum_{v \in N(u)}w_{uv},$
and the variance is:
$\sigma^2 = \P[\hat{Y}(x_u)](1-\P[\hat{Y}(x_u)])\sum_{v \in N(u)}w_{uv}^2.$
Following \citet{zhang2011cotrade}, we approximate
the distribution of $J$ with a normal distribution of mean $\mu$ and variance $\sigma^2$.
Then we can rank examples $x_u$ by the left-sided tail probability for $J_u$ (\emph{lower} is better).
If $J_u$ is much smaller than expected, then the total cut edge weight is much smaller than expected under the null hypothesis.
To select confident data, we sort examples by $J_u$ and choose the top $\beta + t\beta'$ fraction in iteration $t$.
The detailed procedure for confident data selection using the cut statistic is shown in Algorithm \ref{alg:get-conf-data-cs} (supplement).

\subsection{Relabeling.}
Pseudo-labels from previous iterations can either be re-used or thrown out.
If the initial hypothesis has high precision but low coverage, it is typically preferable to re-use the pseudo-labels from previous iterations, since as coverage increases the quality of the pseudo-labels is likely to go down.
On the other hand, if the models being trained are capable of correcting incorrect pseudo-labels, it is preferable to relabel, since this can improve the quality of the training data in each iteration.
We exclusively use the latter, since we found that the pseudo-label accuracy on the covered subset of data often \emph{increased} with more iterations.
The original co-training algorithm \citep{blum1998combining} builds $L$ cumulatively, but subsequent co-training procedures also use relabeling \citep{zhang2011cotrade}.
We discuss relabeling further in Section~\ref{apdx:relabeling}.

\section{Experiments}

\label{sec:experiments}

\paragraph{Datasets.}
We investigate the benefit of co-training on several standard natural language benchmarks, focusing on datasets with a large gap between the best prompt-based methods and fully-supervised learning \citep{wang2019glue, wang2019superglue, Brown2020}.
We use the {\bf RTE} \citep{dagan2005pascal}, {\bf CB} \citep{de2019commitmentbank}, {\bf TREC} \citep{voorhees2000building}, and {\bf BoolQ} \citep{clark-etal-2019-boolq} datasets. 
Full details for these datasets are in Appendix \ref{apdx:training-and-data-details}.
In the partial access setting, we do not evaluate on BoolQ due to the large amount of GPT-3 quota required for labeling.
In the full access setting, we do not evaluate on TREC as T0 was pretrained on TREC.

\paragraph{Training methods, partial access.} 
In the few-shot setting we randomly select $k=4$ training examples from each dataset until the initial label model assigns every pseudo-label at least $\gamma\beta U$ times (i.e., we resample prompts until we can initialize the label model in accordance with the constraint that $\P[Y=y] \ge \gamma$ for all $y$).
While larger $k$ might improve performance, $k=4$ gives a good balance between performance and GPT-3 quota usage. (Indeed, with our 4 one-shot prompts, we are able to beat the GPT-3 32-shot accuracy on CB).

In each co-training iteration, we train the label model over view $\phi_0$ using Adam with learning 1e-4, weight decay 5e-3, and batch size 64 for 40 epochs. We train the DeBERTa-large model over $\phi_1$ for 20 epochs using Adam with learning rate 1e-5, weight decay 0.01, batch size 16. All parameters were frozen except the last language model layer, the pooler, and the linear classification layer. 
In order to avoid indirect label leakage, we did not tune these hyperparameters and instead chose common hyperparameters used for these types of models. For early stopping, each model was evaluated every epoch on a \emph{pseudo}-labeled validation set and the best model checkpoint was chosen based on \emph{balanced accuracy} on the pseudo-labels at the end of each round.
Using the balanced accuracy (average of the recall for each label) avoids collapsing to the majority class even when the pseudo-labels are relatively imbalanced. This validation set was sampled uniformly from the training set to give a training/validation split of 90\%/10\%. 

To determine $\beta,\ \beta', T$ for co-training, we performed a light hyperparameter search based on performance on a \emph{gold}-labeled validation set of 500 examples sampled from the TREC training set.\footnote{Hence our few-shot experiments on TREC are not few-shot in the truest sense of the term.} This resulted in the the following values: initial coverage of $\beta = 0.5$, per-step coverage increase of $\beta' = 0.1$, and total co-training steps $T=5$. We emphasize that this gold validation set was \emph{not} used during any co-training iteration (e.g. for model selection, early stopping, learning rate tuning, etc.) We set the minimum label frequency $\gamma=0.01$ and did not tune this value. We used model confidence to add confident data in view 0 and the cut statistic to add confident data in view 1.\footnote{
This works better than using the cut statistic in both views---since the cut statistic relies heavily on good nearest neighbors, it makes the most sense in a view that already has a good distance function for examples (the pretrained DeBERTa representation).} We used the \texttt{\small [CLS]} token embedding in the last layer of DeBERTa for cut statistic nearest neighbors in view 1.
We used $K = 20$ nearest neighbors for the cut statistic and performed no tuning on this value.

\paragraph{Training methods, full access.}
In the zero-shot setting, for training the soft prompt over view $\phi_0(x)$ we mainly used the hyperparameters suggested by \citet{lester-etal-2021-power}, which were obtained by performing gold soft prompt tuning using T5 on SuperGLUE. We used Adafactor with constant learning rate 0.3, weight decay 1e-5, and batch size 24 for 30000 training steps. For DeBERTa-large, we used the same hyperparameters as in the partial access setting. As in the partial access setting, we used balanced pseudo-label accuracy to select the best model checkpoint at the end of each training round.
We used the cut statistic for confident selection in \emph{both} views, since with T0 we have access to the internal embeddings, unlike with GPT-3.
We used the T0 decoder's contextual embedding for the first decoded token to compute nearest neighbors for the view 0 cut statistic. 
Training details (e.g., $\beta, \beta', T$, etc.) are otherwise exactly the same as in the partial access setting.

During training, the pseudo-label for each example is first mapped to a token that matches the hard prompt (e.g. 0$\to$\texttt{\small True} and 1$\to$\texttt{\small False} for the RTE example above).
These token labels are then mapped to embedding indices using the T0 tokenizer, and the soft prompt is trained via regular sequence-to-sequence training with the maximum likelihood objective.
This is identical to the soft prompt training technique from \citet{lester-etal-2021-power}.

\paragraph{Caveat.} As noted by \citet{perez2021true}, much current work on prompt-based learning does not constitute ``true'' few-shot/zero-shot learning as they often implicitly assume access to a small labeled set to select various model configurations (e.g., prompts and hyperparameters). Insofar as we inherit such configurations  from existing work, our work is similarly not few-shot/zero-shot in the strictest sense, although we tried to minimize such issues by using exactly the same co-training parameters ($\beta, \beta', T, \gamma$) and  model hyperparameters for all datasets. (We also did not perform an extensive tuning of these parameters.) While we are encouraged by the observation that model configurations seem to work well across diverse datasets, investigating co-training in the context of true few-shot/zero-shot learning \cite{schick2021true} is an important avenue for future work.

\input{tables_final}

\paragraph{Baselines.}
For baselines we compare against:
\begin{itemize}[leftmargin=*, topsep=0pt]
    \item \textit{GPT-3 32-shot}: From \citet{Brown2020}. 32 examples combined in one prompt. Uncalibrated.
    \item \textit{Calibrate Before Use}: Performance of CBU using 4-shot prompts (from \citep{zhao2021calibrate}).
    \item \textit{Prompt-based FT}: Our reproduction of the fine-tuning method from \citep{gao2020making}, using 2 labels per class.
    \item \textit{Snorkel on GPT-3 output:} Snorkel generative label model, which aggregates over the four GPT-3 1-shot outputs without using any labeled data. \citep{ratner2016data, ratner2020snorkel}.
    \item \textit{Snorkel + DeBERTA-large}: DeBERTA-large fine-tuned on outputs from Snorkel label model using the same hyperparameters as our co-training methods.
    \item \emph{Label Model (no co-training)}: the label model after initialization with \eqref{eqn:linear-label-model}.
\end{itemize}
The baselines that use (roughly) the same amount of labeled data as our method are shown in the top section of Table~\ref{tbl:lm-results} and Table~\ref{tbl:sp-results}. The bottom sections contain baselines that use more labeled data, including oracle upper-bounds based on full training data.
Training details for the baselines are in Section~\ref{apdx:baseline-details}.

\begin{table*}[!tb]
\centering
\vspace{4mm}
\caption{Zero-shot learning results with T0 as the initial view 0 model and DeBERTa as the second model. We also show the results trained on the full dataset in the bottom two rows. For T0, we take the best prompts from \citet{sanh2021multitask} and replicate their results as exact numbers for each prompt were not provided in the original paper. Standard deviations are not provided in this case as even a single run takes a nontrivial amount of time.}
\vspace{1mm}
\label{tbl:sp-results}
    \begin{tabular}{lcccc}
            \toprule
            Model/Algorithm \hspace{-3mm} & View & RTE & CB  & BoolQ \\
            \midrule
            T0-3B (best) \cite{sanh2021multitask} \hspace{-3mm}  & $\phi_0$ & 68.9 & 66.1 &  59.1 \\
            T0-3B zero-shot (no co-training) \hspace{-3mm}  & $\phi_0$ & 68.9 & 58.9 &  56.4 \\
        T0-3B soft prompt + \emph{co-training} \hspace{-3mm}   & $\phi_0$ & 87.0 & 67.9 & 49.1 \\
         DeBERTa-large + \emph{co-training} \hspace{-3mm}  & $\phi_1$ & 86.3 & 67.9 & 48.9 \\
         \midrule
                  T0-3B soft prompt on full train & $\phi_0$ & 90.6 & 80.4 \hspace{-2mm}  &  86.9 \\
            DeBERTa-large on full train & $\phi_1$ & 93.3 & 95.2 &  86.1 \\
            \bottomrule
    \end{tabular}
        \vspace{1.5mm}
\end{table*}

\subsection{Results}

Table \ref{tbl:lm-results} shows the results for the partial access setting, where we co-train the label model, which calibrates and combines multiple GPT-3 outputs, with a smaller pretrained model (DeBERTa-large).
For view 0, our co-trained label model (\emph{Label Model + co-training}) improves over the initial label model (\emph{Label Model before co-training}) and the average performance of GPT-3 4-shot before (\emph{GPT-3 4-shot}) and after (\emph{Calibrate Before Use}) calibration. 
It also improves over \emph{Snorkel on GPT-3}, which, like our method, uses unlabeled data to combine the outputs of our four 1-shot prompts.
For CB, the co-trained label model outperforms \emph{GPT-3 32-shot} despite only using 4 labeled examples.
This suggests that using unlabeled data to learn to ensemble $k$ 1-shot prompts can be more label-efficient than putting all $k$ labeled examples in one prompt.
For TREC and CB, the co-trained label model also outperforms prompt-based fine-tuning (\emph{Prompt-based FT} \citep{gao2020making}) with the same amount of labeled data (Prompt-based FT also uses a gold-labeled validation set of $k$ examples per class, whereas our method only uses a pseudo-labeled validation set).
For RTE and CB, we nearly match the fully-supervised performance on view 0 (\emph{Label Model on full train}), suggesting that co-training is able to extract nearly all of the signal from the GPT-3 probabilities in these cases without using any extra labeled data.

For view 1, the co-trained DeBERTa-large model outperforms all of the baselines that use the same amount of label information.
For RTE and TREC, it outperforms Prompt-based FT even when the latter uses 4x (for RTE) and 12x (for TREC) the number of labeled examples (\emph{Prompt-based FT with 8 labels per class}).
This suggests that the pseudo-labels provided by (a learned ensemble of) prompts are an effective training signal for smaller models.

Table \ref{tbl:sp-results} shows the results for the full access setting with T0.
For RTE and CB, co-training improves on the performance of the initial zero-shot prompt (\emph{T0-3B zero-shot (no co-training)}).
For RTE, the co-trained view 0 and view 1 models nearly match the performance of their fully-supervised counterparts.
The difference in co-training performance on RTE in Table \ref{tbl:lm-results} and Table \ref{tbl:sp-results} shows the benefit of having full access to $h_0$.
Since we can introspect the prompt model, we can use the cut statistic in both views. In the first step of co-training, the cut statistic on view 0 selects confident data with 90\%-accurate pseudo-labels.
The confident data selection on CB is similarly good: in the first co-training step, the cut statistic selects pseudo-labels with 89\% accuracy.
The pseudo-labels extracted by the view 1 model after the first step of co-training ($L_1^0$) are 98\% accurate, so after training on the initial pseudo-labels, the view 1 model is able to select a very high quality training set at coverage $\beta=0.5$.
However, the CB performance is \emph{worse} than in Table \ref{tbl:lm-results} despite the strong initial signal in $L_0^0$ and the near-perfect training data in $L_1^0$.
Similarly, for BoolQ, co-training makes the soft prompt \emph{worse} than the initial zero-shot model. 
We explore the reasons behind this below. %

\subsection{When (and how) does co-training work?}
\begin{figure}[tb]
    \centering
    \begin{subfigure}{0.5\linewidth}
        \centering
        \includegraphics[width=\linewidth]{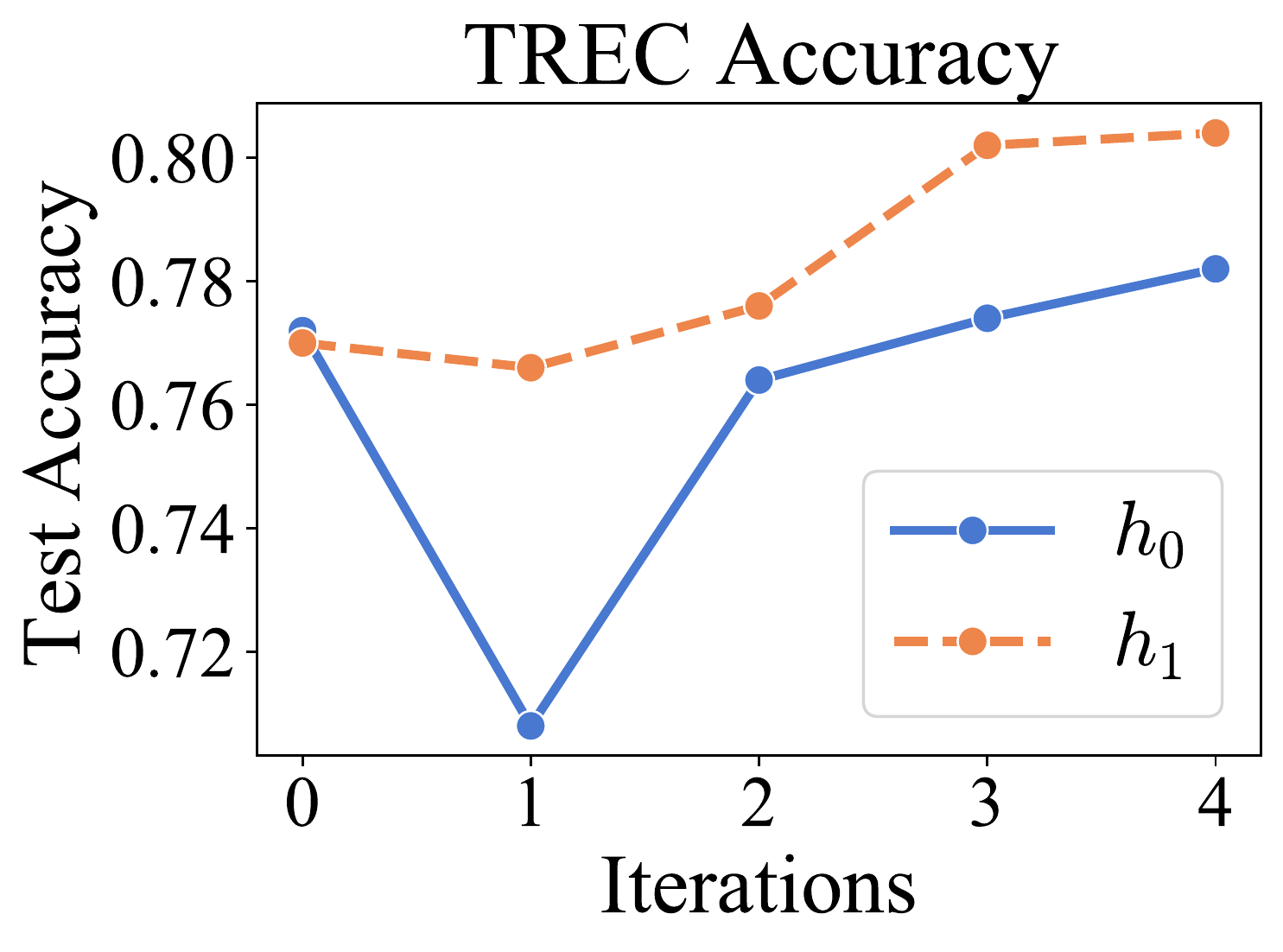}
    \end{subfigure}\hfill%
    \begin{subfigure}{0.5\linewidth}
        \centering
        \includegraphics[width=\linewidth]{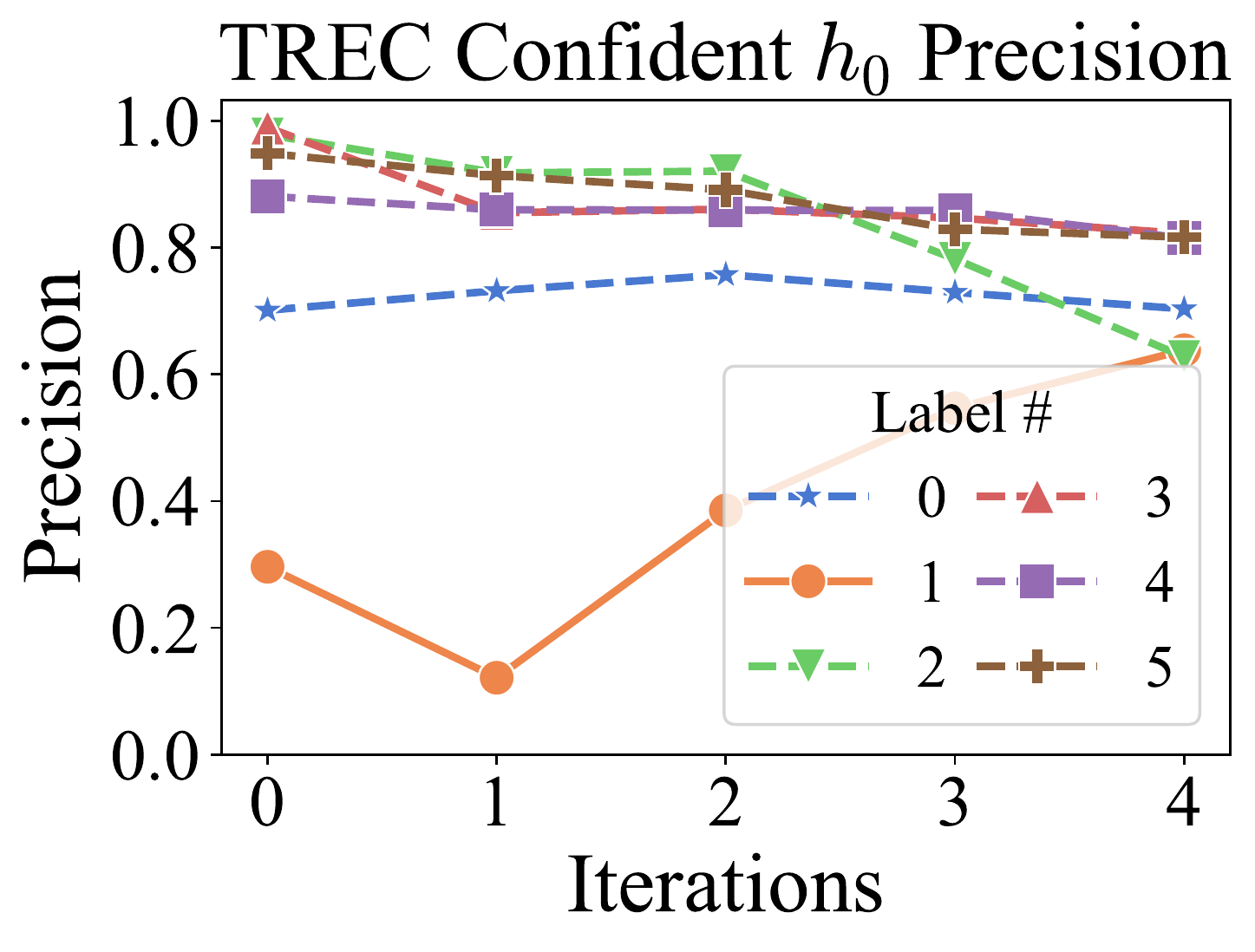}
    \end{subfigure}%
    \caption{Partial access setting, TREC. Left: Test accuracy vs co-training iteration for the label model $h_0$ and the DeBERTa model $h_1$. Right: precision per label vs co-training iteration, $h_0$.}
    \label{fig:cotrain-evolution}
\end{figure}

Figure~\ref{fig:cotrain-evolution}~(left) shows the evolution of the test accuracy for $h_0$ and $h_1$ over co-training iterations on TREC (from Table~\ref{tbl:lm-results}).
Multiple rounds of co-training increase performance for both views as the models become more precise and the coverage increases.
Figure~\ref{fig:cotrain-evolution}~(right) shows the precision of the \emph{confident} data extracted by $h_0$ for each iteration of co-training, broken down by label.
This figure shows two distinct phenomena.
For most labels, precision decreases as coverage goes up, as we expect from usual co-training theory \citep[see e.g.][]{balcan2005co}.
However, for label 1, precision actually \emph{increases} over iterations.
Model $h_1$ (DeBERTa) is able to select new confident data for label 1 that is \emph{better} than the weak labels used to train it, which improves the $h_0$ precision for label 1 in subsequent iterations.
For example, in iteration 2, $h_0$'s confident precision for label 1 is 0.39, but after $h_1$ is trained on that data, it proposes new confident data for label 1 with precision 0.58 (not shown in Figure~\ref{fig:cotrain-evolution}).
Pseudo-label correction is one of the benefits of having two complementary models (though it can also happen with a single model with appropriate regularization \citep{wei2020theoretical}).

\begin{figure}[tb]
    \centering
    \begin{subfigure}{0.5\linewidth}
        \centering
        \includegraphics[width=\linewidth]{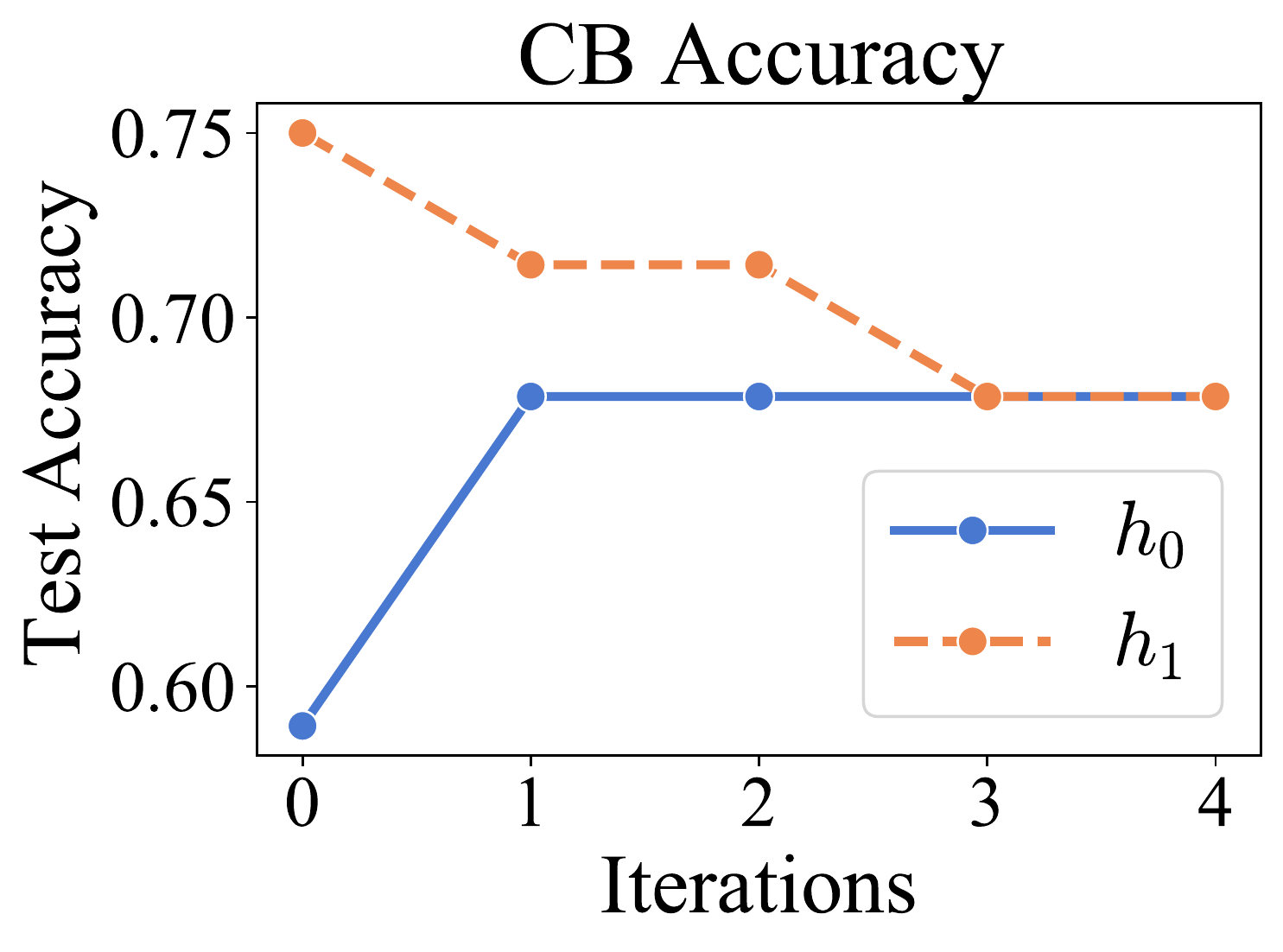}
    \end{subfigure}\hfill%
    \begin{subfigure}{0.48\linewidth}
        \centering
        \includegraphics[width=\linewidth]{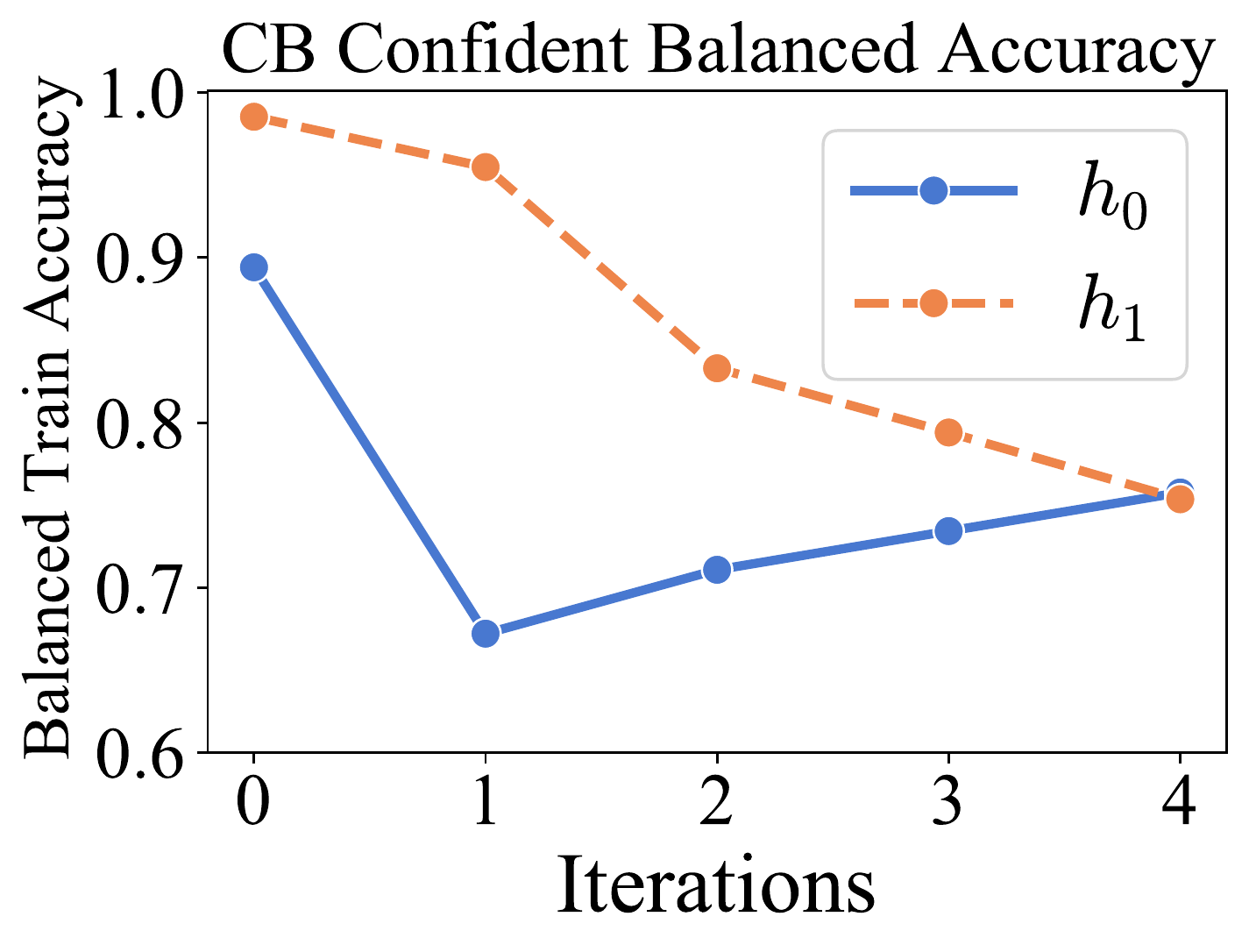}
    \end{subfigure}%
    \caption{Full access setting, CB. Left: Test accuracy vs co-training iteration for T0-3B ($h_0$) and the DeBERTa model ($h_1$). Right: balanced accuracy of confident pseudo-labels extracted from T0-3B ($h_0$) and DeBERTa ($h_1$).}
    \label{fig:cotrain-cb}
\end{figure}

Figure \ref{fig:cotrain-cb} shows what can happen when $h_0$ and $h_1$ are not complementary enough.
The left display shows the accuracy of each model over co-training iterations.
The right display shows the balanced accuracy of the confident data extracted from each model.
In the first co-training step, $h_1$ greatly improves over the initial $h_0$, and selects extremely accurate confident data (nearly 100\% accurate) at coverage $\beta=0.5$.
This improves the performance of the soft prompt for the next iteration ($h_0$, left, iteration 1), but the confident balanced accuracy of $h_0$ sharply decreases.
Inspecting the training of $h_0$ on $L_1^0$, the soft prompting procedure appears to have overfit to the pseudo-labels on $L_1^0$.
Due to the small size of CB (250 training examples), coverage $\beta=0.5$ and our 90/10 train/val split gives only 112 data points for training $h_0$ in the first iteration, but the soft prompt is a very flexible hypothesis class.
This overfitting causes the accuracy of confident data to decrease, which in turn degrades the performance of $h_0$, and eventually the two models converge to almost identical predictors.
This suggests that CB does not have enough unlabeled data for co-training to perform well with T0, at least when $\beta=0.5$.
Using larger initial coverage (e.g. $\beta=1.0$, $\beta'=0$) or a less flexible hypothesis class for $h_0$ might improve performance.

Finally, Table \ref{tbl:sp-results} showed that co-training \emph{decreased} performance on BoolQ even though the initial soft prompt seemed to have reasonably strong signal (56.4\% accuracy).
However, the finer-grained statistics are less promising. After the first training iteration, with $\beta=0.5$, $h_1$ assigns pseudolabel 0 to 3270 training examples with precision 0.4 and pseudolabel 1 to 1848 examples with precision 0.66.
This means the ``total noise'' $\eta = \P[\hat{Y}=1|Y=0] + \P[\hat{Y}=0|Y=1]$ in the pseudo-labels is 0.93. At the beginning of iteration $t=1$, when $\tilde{\beta}=0.6$, the total noise in the confident data assigned by $h_0$ is 0.98.
For comparison, the total noise for the initial $h_0$ on CB is 0.21.
Even under ideal conditions on $\phi_0$ and $\phi_1$, $\eta < 1$ is required for learning to work, and the sample complexity depends on $1/(1-\eta)$ \citep{blum1998combining}.
Unfortunately, the same issue persists on BoolQ for different $\beta$ values.
The negative result on BoolQ suggests that the initialization for $h_0$ needs to have less total noise.
A different prompt or a better initial hypothesis (e.g., full T0 instead of T0-3B) could be more amenable to co-training.

\section{Conclusion}

Our results indicate that using unlabeled data to co-train a prompted model with a smaller model can boost the performance of prompt-based learning on few-shot and zero-shot classification tasks.
As a side effect, this procedure also produces a smaller performant model on the task of interest, distilling and refining the knowledge in the large prompted model.
Using two complementary models and views allows the models to learn from each other despite training on partially incorrect pseudo-labels.
We showed that the benefit of co-training is limited when the initial signal provided by the prompted model is too noisy (BoolQ, full access), when there is not enough unlabeled data to obtain good (pseudo-label) generalization performance (CB, full access), and when there is a large gap in fully-supervised accuracy on view 0 and view 1 (RTE, partial vs full access).
Developing methods to overcome these limitations in the context of prompting is an interesting direction for future work.

\section*{Acknowledgments}
DS and HL were partially supported by NSF AiTF award CCF-1723344.
MA was supported by the Takeda Fellowship.
Thanks to Dr. Steven Horng of Beth Israel Deaconess Medical Center for providing access to an NVIDIA DGX machine \citep{horng2022machine}, and thanks to NVIDIA Corporation for their donation of two NVIDIA A100 GPUs.
Thanks to OpenAI and AI21 for providing quota to access their \texttt{davinci} and Jurassic-Jumbo models (respectively).
Finally, thanks to Rebecca Boiarsky and Catherine Wong for their feedback on drafts of this paper and to Aravindan Vijayaraghavan for helpful discussions on co-training theory.
\clearpage
\bibliography{custom}
\bibliographystyle{arxiv}
\clearpage
\appendix
\input{appendix}
\end{document}

%% file: notation.tex
\DeclareMathOperator*{\argmax}{\text{argmax}}

\DeclareMathOperator{\softmax}{softmax}
\DeclareMathOperator{\ReLU}{ReLU}
\DeclareMathOperator{\diag}{Diag}
\DeclareMathOperator{\gpt}{GPT3}

\renewcommand{\P}{\mathbb{P}}

\newtheorem*{proposition*}{Proposition}

\newtheorem*{theorem*}{Theorem}

%% file: tables_final.tex
\begin{table*}[!tb]
\centering
   \vspace{-3mm}
\caption{Few-shot learning results. (Top) Results against various baselines that use exactly 4 labels per dataset (except for Prompt-based FT, which uses 8 labels for CB and 12 labels for TREC, since this approach uses labels at the ``per-class'' level). GPT-3 and CBU results are copied from \citet{zhao2021calibrate}, while we train our own Prompt-based FT \cite{gao2020making} and Snorkel \cite{ratner2020snorkel} models. (Bottom) Results from baselines trained on more data (for reference only).  Standard deviation (when applicable) numbers are given by 4 runs over prompts (GPT-3, CBU) or random seeds (Snorkel, Prompt-based FT, Co-training). $\dagger$ rows show accuracy on the private SuperGLUE test set. Otherwise, the accuracies are on the public SuperGLUE validation sets, which we treated as a test set.}
\label{tbl:lm-results}
\vspace{1mm}
    \begin{tabular}{lcllll}
            \toprule
            Model & View & RTE (2-class) & CB (3-class) & TREC (6-class) \\
            \midrule
     
            GPT-3 4-shot (from \citet{zhao2021calibrate}) & * & 58.7 (11.9) & 45.2 (19.4) & 60.2 (7.6)  \\
            Calibrate Before Use (CBU) \cite{zhao2021calibrate} & * & 60.4 (8.1) & 60.7 (6.7) & 69.7 (1.4)\\

            Prompt-based FT \citep{gao2020making} & * & 52.8 (0.9) & 84.4 (3.2) & 54.8 (2.9)\\
            Snorkel on GPT-3 \citep{ratner2020snorkel}  & $\phi_0$ &  59.6 (0.0) & 70.2 (0.8) &  65.2 (0.0)\\
            Snorkel on GPT-3 + DeBERTa-large & $\phi_1$ &67.2 (0.5) & 81.6 (2.2) &  63.3 (0.4)\\

            \text{Label Model (no co-training)} & $\phi_0$ &  62.8 & 76.8 & 77.2\\
           \text{Label Model + \emph{co-training}}& $\phi_0$ &  64.9 (1.1) & 83.5 (2.3) &  78.3 (1.2)\\
           \text{DeBERTa-large + \emph{co-training}} & $\phi_1$ &  {\bf 67.4} (2.3) & {\bf 86.2 } (3.2) & {\bf 80.6} (1.1)\\
\midrule
            Label Model  on full train &  $\phi_0$ & 67.8 (0.5)  & 82.7 (0.8) & 91.9 (1.1)  \\
            DeBERTa-large on full train &  $\phi_1$ & 93.3 & 95.2 & 96.7  \\
                        GPT-3 32-shot$^{\dagger}$ \cite{Brown2020} &   * &69.0 & 75.6 &  * \\
          Prompt-based FT with 4 labels per class & * & 48.3 (3.3) & 84.4 (4.4) & 58.8 (5.9)\\
           Prompt-based FT with 8 labels per class  & * & 56.1 (3.2) & {87.5} (0.0) & 80.0 (4.8)\\
        \bottomrule
    \end{tabular}
   \vspace{-3mm}
\end{table*}

%% file: appendix.tex
\section{Algorithm details}
\label{apdx:algorithm-details}
\subsection{Relabeling}
\label{apdx:relabeling}
Pseudolabels from previous iterations can either be re-used or thrown out.
If the initial hypothesis has high precision but low coverage, it is typically preferable to re-use the pseudolabels from previous iterations, since as coverage increases the quality of the pseudolabels is likely to go down.
On the other hand, if the models being trained are capable of correcting incorrect pseudolabels, it is preferable to relabel, since this can improve the quality of the training data in each iteration.

In iteration $t$ of co-training, we extract a pseudolabeled dataset $L^t_i$ from model $h_i$ and use it to train model $h_{1-i}$.
Let $\mathcal{N} \subset [U]$ be a set of indices that correspond to the data points confidently pseudolabeled by model $h_i$ in this iteration (according to either the model confidence or cut statistic rankings).
Define $S = \{(x_n, \hat{y}_n) : n\in \mathcal{N}\}$ as the set of these points together with their pseudolabels $\hat{y}_n := \argmax h_i(\phi_i(x_n))$.
Let $L^{t-1}_i = \{(x_n, \tilde{y}_n)\}$ be the confident data used in the \emph{previous} iteration to train model $h_{1-i}$.
For $x_n$ that appear in $L^{t-1}_i$ but where $n\in \mathcal{N}$, we have a choice to make: do we use the old pseudolabel $\tilde{y}_n$, or the new one $\hat{y}_n$?
These need not agree, since $h_i$ has been updated.

Let $S' = \{(x_n, \hat{y}_n) \in S\ |\ \neg\exists y : (x_n, y) \in L^{t-1}_i\}$ be the set of \emph{newly} pseudolabeled examples---points that do not appear in $L^{t-1}_i$ with any pseudolabel.
If we choose to re-use the pseudolabels from the previous iteration, the \texttt{GetConfData} update is:
\[
L^t_i \gets L^{t-1}_{i} \cup S'
\]
On the other hand, if we throw out the previously pseudolabeled data, the update is simply:
\[
L^t_i\gets S
\]
We exclusively use the latter, since our models can learn to correct bad initial labels (see Figure \ref{fig:cotrain-evolution}, Label 1).
We found that the pseudolabel accuracy on the covered subset of data often \emph{increased} with more iterations.
This relabeling technique is different from the original cotraining algorithm \citep{blum1998combining}, which builds $L$ cumulatively, but subsequent cotraining procedures also use relabeling \citep{zhang2011cotrade}.

\subsection{Warm starting}
Instead of warm-starting the models $h_0$ and $h_1$ (initializing them from the output of the previous iteration), we initialize them from scratch each co-training iteration to reduce the effect of a ``bad'' training iteration and so that we can use the same training hyperparameters for every iteration. This takes advantage of the fact that there exists a robust set of initial hyperparameters that have been shown to work well for fine-tuning large language models. However, further exploration of warm-starting is an interesting direction for future work, since it may yield significant reduction in the computational burden of co-training.

\subsection{Confident data selection}
Algorithm \ref{alg:get-conf-data-mc} shows how to select confident data using model confidence, and
Algorithm \ref{alg:get-conf-data-cs} shows how to select confident data using the cut statistic.
As mentioned in the main text, $\phi_0$ and $\phi_1$ themselves needn't be the representations used to compute nearest neighbors for the cut statistic.
For example, $\phi_0(x)$ for T0 is the non-contextual T0 input embedding of $x$.
Instead of computing nearest neighbors in this view, we use the \emph{contextual} embedding from much later in the T0 model: the final decoder embedding of the first decoded token.
This is a function of both $\phi_0(x)$ and the current hypothesis $h_0$.
Because the embeddings are contextual, this representation has better nearest neighbors than $\phi_0(x)$; because it also takes $h_0$ into account, these neighbors are adapted to the current task.
Similarly, for DeBERTa-large in view 1, we use the  \texttt{[CLS]} token embedding in the \emph{last} layer of the DeBERTa representation rather than the penultimate layer, since this layer has been adapted to the task of interest by the time we select confident data.
\paragraph{Validation dataset}
We use a pseudolabeled validation set to perform model selection during the co-training iterations.
Since the confident-data-selection methods can pick out the most precisely pseudolabeled examples (w.r.t. the true label), we also use them to select a \emph{confident} validation set from the larger val set for each \texttt{Train} step. In particular, when training model $h_i$, we use model $h_{1-i}$ to select a $\tilde{\beta} = \beta + t\beta'$ confident fraction of full validation data in each step (the same fraction used for the confident training set).
This allows us to use a more precise validation set for model selection.

\subsection{Full algorithms}
The detailed algorithms for co-training in the partial access setting and full access setting are shown in Algorithms \ref{alg:cotrain-gpt3-detailed} and \ref{alg:cotrain-t0-detailed}, respectively.
Algorithm \ref{alg:cotrain-gpt3-detailed} uses model confidence for view 0 and cut statistic for view 1.
Algorithm \ref{alg:cotrain-t0-detailed} uses cut statistic for both views.
The detailed procedures for model confidence and the cut statistic are shown in Algorithms \ref{alg:get-conf-data-mc} and \ref{alg:get-conf-data-cs}, respectively.
As mentioned in the previous section, the view 1 cut statistic uses the \texttt{[CLS]} token embedding in the last layer of $h_1$.
In the full access case, the view 0 cut statistic uses the T0 decoder's hidden state for the first decoded token.

\clearpage
\section{Training and dataset details}
\label{apdx:training-and-data-details}
\subsection{Datasets}
\begin{itemize}[leftmargin=*]
\setlength\itemsep{0em}
    \item \textbf{RTE} \citep{dagan2005pascal}: Binary textual entailment, 2490 training examples, 277 validation examples (our test set).
    \[
        \P[Y] = (0.5, 0.5)
    \]
    \item \textbf{CB} \citep{de2019commitmentbank}: Ternary textual entailment, 250 training examples, 56 validation examples (our test set).
    \[
        \P[Y] = (0.41, 0.5, 0.09)
    \]
    \item \textbf{WiC} \citep{pilehvar2018wic}: Binary word sense disambiguation, 5428 training examples, 638 validation examples (our test set).
    \[
    \P[Y] = (0.5, 0.5)
    \]
    \item \textbf{TREC} \citep{voorhees2000building}: 6-way question classification, 5452 training examples, 500 test examples. Label balance: 
    \[\P[Y] = (0.2131, 0.2293, 0.0158, 0.2243, 0.1643, 0.1532)\]
    \item \textbf{BoolQ} \citep{clark-etal-2019-boolq}: Binary reading comprehension, 9427 training examples, 3270 validation examples (our test).
    \[
    \P[Y] = (0.38, 0.62)
    \]
\end{itemize}
\subsection{Training details}
\label{apdx:baseline-details}
\paragraph{Prompt-based FT.}
We fine-tuned the MLM-pretrained RoBERTa-large model using Adam for 1000 steps with batch size 16, learning rate 1e-5 and weight decay 0.01.
We sampled a validation set the same size as the training set while ensuring that the validation set also had an equal number of examples per class.
This small validation set was used to select the best model checkpoint in each run and the test results were averaged over four random seeds.
This is similar to the ``no $\mathcal{D}_{dev}$'' setting in \citet{gao2020making} in that we didn't use the small validation set for hyperparameter tuning---we used the same hyperparameters as the ``no $\mathcal{D}_{dev}$'' setting. However, we still allow the method to use the labeled validation set for model selection.
We used the same prompt templates as  \citet{gao2020making}. For RTE, the label words were \texttt{Yes}, \texttt{No}. For CB, the label words were \texttt{Yes}, \texttt{No}, \texttt{Maybe}.
For TREC, the label words were \texttt{Description}, \texttt{Entity}, \texttt{Abbreviation}, \texttt{Person}, \texttt{Number}, \texttt{Location}.

\paragraph{Calibrate Before Use (CBU).}
For $x_{cf}$, we followed \citet{zhao2021calibrate} and used ``N/A'', the empty string, and ``[MASK]''.
We obtained the GPT-3 outputs for each of these $x_{cf}$'s, renormalized the outputs over the label tokens, averaged the re-normalized outputs across the three $x_{cf}$'s, and used the \emph{average} result as the scaling factor for $W^{(i)}$. 
This is identical to \citet{zhao2021calibrate}.

\paragraph{Co-training.}
Following RoBERTa and DeBERTa, we used an MNLI-pretrained checkpoint for RTE and CB (\texttt{microsoft/deberta-large-mnli} on HuggingFaceHub).
Otherwise, we used DeBERTa-large (\texttt{microsoft/deberta-large}).
We did not experiment with DeBERTa V2 or V3.

\subsection{Soft prompt encoding}
\label{apdx:sp-encoding}
As detailed in Section \ref{sec:co-training-desc}, we combine hard prompt encoding with soft prompting.
That is, we format the input using a hard prompt, and combine this \emph{formatted} input embedding with the soft prompt matrix.
This differs from the usual soft prompting setup \citep{li-liang-2021-prefix, lester-etal-2021-power}, where the input is encoded more neutrally, without a natural-language hard prompt:
\vspace{1mm}
\begin{displayquote}
\texttt{sentence1: \{\{$x$.premise\}\}\\} \texttt{sentence2: \{\{$x$.hypothesis\}\}}
\end{displayquote}
\vspace{1mm}
\emph{A priori}, this difference in input encoding could affect the performance of soft prompt tuning and the zero-shot performance of the initial prompted model. However, the full-training-dataset soft-prompt tuning baseline in Table \ref{tbl:sp-results} (\emph{T0 soft prompts on full training set}) uses our hard prompt encoding + soft prompting, and it matches fully fine-tuned DeBERTa-large.
This suggests that the accuracy loss from choosing a hard prompt (at least for the prompts that we chose) is minimal. 

Using the hard prompt encoding might improve the label efficiency of soft prompt tuning, since the soft prompt parameters can focus on ``fixing up'' the given hard prompt instead of learning a prompt-like embedding from scratch.
On the other hand, if the hard prompt performs poorly, the hard prompt encoding might put an unnecessary upper limit on the soft prompt tuning performance, since the soft prompt may not be able to ``undo'' the hard prompt performance. 
An in-depth comparison between the neutral encoding from the traditional soft-prompting setup and the hard prompt + soft prompt encoding we propose is an interesting direction for future work.

\subsection{Hardware}
All models were trained on two NVIDIA A100 80Gb GPUs using PyTorch and the Transformers library \citep{wolf-etal-2020-transformers}.
For the partial access setting, a full run of $T=5$ co-training iterations with DeBERTa-large takes roughly two hours on this hardware. For the full access setting, a full run of $T=5$ co-training iterations with T0-3B and DeBERTa-large takes roughly 40 hours.
\vspace{4mm}
\input{appendix_deepdive}

\clearpage
\onecolumn
\section{Prompts}
Here we list the prompts used for our experiments, largely taken from \citet{sanh2021multitask}.

\subsection{Partial Access Setting}
\textbf{RTE}\\
{\small
\texttt{\{example.premise\} \\
Question:\{example.hypothesis\} True, False, or Unknown?\\
answer: \{example.answer\} \\
\{premise\} \\
Question: \{hypothesis\} True, False, or Unknown? \\
answer:}}

\noindent\textbf{CB}\\
{\small 
\texttt{premise: \{example.premise\} \\
hypothesis:  \{example.hypothesis\}\\
Does the premise imply the hypothesis? Yes, No, or Neither? \\
answer: \{example.answer\}\\
premise: \{premise\}\\
hypothesis: \{hypothesis\}\\
Does the premise imply the hypothesis? Yes, No, or Neither?\\
answer:}}

\noindent\textbf{TREC}\\ 
\texttt{Classify the questions based on whether their answer type is Unknown, Number, Location, Person, Description, Entity, or Abbreviation.\\
Question: \{example\_question\} \\ 
Answer Type: \{example\_type\} \\
Question: \{question\} \\
Answer Type: }

\subsection{Full Access Setting}
\noindent\textbf{RTE}\\
\texttt{\{premise\}\\
Question: \{hypothesis\} True, False, or Unknown?}

\noindent\textbf{CB}\\
\texttt{\{premise\} \\
Question: \{hypothesis\} True, False, or Neither?}

\noindent\textbf{BoolQ}\\ 
\texttt{Text: \{passage\} \\
Answer the following yes/no question: \{question\}? Yes or no?}

\clearpage
\twocolumn
\begin{figure*}
\begin{minipage}[t]{0.48\textwidth}
\begin{algorithm}[H]
  \caption{\texttt{GetConfDataMC}}
  \label{alg:get-conf-data-mc}
\begin{algorithmic}
\INPUT $\{x_n\}_{n=1}^U$ unlabeled examples
\INPUT model $h_i$, view $\phi_i$
\INPUT coverage fraction $\tilde{\beta}$
\INPUT minimum class percentage $\gamma$
\STATE {\color{gray}// compute pseudolabel and score for each example}
 \FOR{$n \mbox{ in } \{1,\ldots, U\}$}
    \STATE $o_n = h_i(\phi_i(x_n))$\ \ \ \  (note $o_n \in \mathbb{R}^{numlabels}$)
    \STATE $\hat{y}_n \gets \argmax_l o_{nl}$
    \STATE $s_n \gets \max_l o_{nl}$
  \ENDFOR
\STATE $L \gets \emptyset$
\STATE
\STATE {\color{gray} // first, select top $\gamma$\% for each class by score}
\STATE $ms \gets \lfloor \gamma \tilde\beta U\rfloor$ {\color{gray} // min num points to select}
\FOR{$l \mbox{ in } \{1,\ldots, numlabels\}$}  
    \STATE $\mathcal{I}_l = \{(x_n, \hat{y}_n, s_n) : \hat{y}_n = l\}$
    \STATE {\color{gray}// sort $\mathcal{I}_l$ by score $s_n$ (ascending)}
    \STATE $S_l \gets $\texttt{Sort}($\mathcal{I}_l$, key=lambda q: q[2])
    \STATE {\color{gray} // add top $\alpha$\% to $L$ (read off end of $S_l$)}
    \STATE $L \gets L \cup S_l$\texttt{[-$ms$:]}
\ENDFOR
\STATE
\STATE {\color{gray}// now select the rest of the points}
\STATE $rs \gets \lceil \tilde\beta U\rceil - |L|$ {\color{gray}// num remaining points to select}
\STATE $\mathcal{I} = \{(x_n,\hat{y}_n, s_n)\}_{n=1}^U$
\STATE $\mathcal{I} \gets \mathcal{I} \setminus L$ {\color{gray}// don't select twice}
\STATE $S \gets $\texttt{Sort}($\mathcal{I}$, key=lambda q: q[2])
\STATE $L \gets L \cup S$\texttt{[-$rs$:]}
\STATE
\STATE {\color{gray}// chop off score and return point + pseudolabel}
\STATE $L \gets \{(x_n, \hat{y}_n) | \exists s_n : (x_n, \hat{y}_n, s_n) \in L\} $
\STATE {\bfseries return} $L$
\end{algorithmic}
\end{algorithm}
\end{minipage}\hfill%
\begin{minipage}[t]{0.48\textwidth}
\begin{algorithm}[H]
  \caption{\texttt{GetConfDataCS}}
  \label{alg:get-conf-data-cs}
\begin{algorithmic}
  \INPUT $\{x_n\}_{n=1}^U$ unlabeled examples
  \INPUT model $h_i$, view $\phi_i$
  \INPUT coverage fraction $\tilde{\beta}$  
  \STATE {\color{gray}// compute pseudolabel and repr. for each example}
  \FOR{$n \mbox{ in } \{1,\ldots, U\}$}
    \STATE $o_n = h_i(\phi_i(x_n))$\ \ \ \  (note $o_n \in \mathbb{R}^{numlabels}$)
    \STATE $\hat{y}_n \gets \argmax_l o_{nl}$
  \ENDFOR
\STATE $L \gets \emptyset$
\FOR{$y \mbox{ in } \{1,\ldots, numlabels\}$}
    \STATE $\hat{P}_y = |\{n : \hat{y}_n = y\}|/U$
\ENDFOR
\STATE
\STATE {\color{gray}// compute 20 nearest neighbors for each ex.}
  \FOR{$u \mbox{ in } \{1,\ldots, U\}$}
    \STATE $N(u) = \mathrm{NN}_{20}(\phi_i(x_u), \{\phi_i(x_{v})\}_{v=1}^U)$
    \FOR{$v \mbox{ in } N(u)$}
        \STATE $w_{uv} = 1/(1+\Vert\phi(x_u) - \phi(x_v)\Vert_2)$
        \STATE $I_{uv} = \mathbb{I}[\hat{y}_u \ne \hat{y}_v]$
    \ENDFOR 
    \STATE {\color{gray}// now compute cut statistic}
    \STATE $J_u = \sum_{v\in N(u)}w_{uv}I_{uv}$
    \STATE $\mu_u = (1-\hat{P}_{\hat{y}_u})\sum_{v \in N(u)}w_{uv}$
    \STATE $\sigma^2 = \hat{P}_{\hat{y}_u}(1-\hat{P}_{\hat{y}_u})\sum_{v \in N(u)}w_{uv}^2$
    \STATE $s_u = \frac{J_u - \mu_u}{\sigma}$
  \ENDFOR
\STATE {\color{gray}// now sort by statistic and return top data}  
\STATE $\mathcal{I} \gets \{(x_n, \hat{y}_n, s_n)\}_{n=1}^U$
\STATE $S \gets $\texttt{Sort}($\mathcal{I}$, key=lambda q: q[2])
\STATE $ns = \lfloor \tilde\beta U\rfloor$
\STATE $L \gets S$\texttt{[:ns]}
\STATE $L \gets \{(x_n, \hat{y}_n) | \exists s_n : (x_n, \hat{y}_n, s_n) \in L\} $
\STATE {\bfseries return} $L$
\end{algorithmic}
\end{algorithm}
\end{minipage}
\end{figure*}

\begin{figure*}
\begin{minipage}[t]{0.48\textwidth}
\begin{algorithm}[H]
  \caption{Co-training algorithm (detailed, GPT-3)}
  \label{alg:cotrain-gpt3-detailed}
  \begin{algorithmic}
  \INPUT $\{(x_j, y_j)\}_{j=1}^k$ initial labeled examples
  \INPUT $\mathcal{U} = \{x_n\}_{n=1}^U$ unlabeled examples
  \INPUT initial coverage $\beta$, coverage increase $\beta'$
  \INPUT minimum percentage per class $\gamma$
  \STATE {\color{gray}// build view 0 for unlabeled examples}
  \FOR{$n \mbox{ in } \{1,\ldots, U\}$}
  \FOR{$j \mbox{ in } \{1,\ldots, k\}$}
  \STATE $\phi_0^{(j)}(x_n) \gets \gpt({x_n, (x_j, y_j)})$
  \ENDFOR
  \STATE $\phi_0(x_n) \gets \left(\phi_0^{(1)}; \ldots;\phi_0^{(k)}\right)$
  \ENDFOR
  \STATE 
  \STATE {\color{gray}// build view 1 for unlabeled examples}
  \FOR{$n \mbox{ in } \{1,\ldots, U\}$}
  \STATE {\color{gray}// extract pre-trained DeBERTa representation for $x_n$}
  \STATE $\phi_1(x_n) \gets \mathrm{DeBERTa}(x_n)$
  \ENDFOR
  \STATE
  \STATE {\color{gray}// initialize $h_0$ according to \eqref{eqn:linear-label-model}}
  \FOR{$j \mbox{ in } \{1,\ldots, k\}$}
    \STATE {\color{gray}// get GPT-3 outputs on content-free input}
    \STATE $\phi_0^{(j)}(x_{cf}) \gets \gpt(x_{cf}, (x_j, y_j))$
    \STATE $W^{(j)} \gets \diag\left(\frac{1}{\phi_0^{(j)}(x_{cf})}\right)$
  \ENDFOR
  \STATE $W \gets \{W^{(j)}\}_{j=1}^k$
  \STATE $\alpha \gets {\bf 1}$
  \STATE $h_0 \gets h_0(\ \cdot\ ; W, \alpha)$
  \STATE
  \STATE {\color{gray}// co-training loop}
  \FOR{$t \mbox{ in } \{0,\ldots, T-1\}$}
      \STATE $\tilde\beta \gets \beta + t\beta'$
      \STATE $L_0^t \gets$ \texttt{GetConfDataMC}($\mathcal{U}, h_0, \phi_0, \tilde\beta, \gamma$)
      \STATE $h_{1} \gets$ \texttt{Train}($\phi_{1}, L_0^t$)
      \STATE $L^t_1 \gets$ \texttt{GetConfDataCS}($\mathcal{U}, h_{1}, \phi_{1}, \tilde\beta$)
      \STATE $h_{0} \gets$ \texttt{Train}($\phi_{0}, L_1^t$)
  \ENDFOR
  \STATE {\bfseries return} $(h_0, h_1)$
\end{algorithmic}
\end{algorithm}
\end{minipage}\hfill%
\begin{minipage}[t]{0.48\textwidth}
\begin{algorithm}[H]
  \caption{Co-training algorithm (detailed, T0)}
  \label{alg:cotrain-t0-detailed}
  \begin{algorithmic}
  \INPUT $\mathcal{U} = \{x_n\}_{n=1}^U$ unlabeled examples
  \INPUT initial coverage $\beta$, coverage increase $\beta'$
  \STATE {\color{gray}// format the input with a hard prompt template $P$,}
  \STATE {\color{gray}// then get T0 embedding to build view $\phi_0$}
  \FOR{$n \mbox{ in } \{1,\ldots, U\}$}
  \STATE $\tilde{x}_n \gets P(x_n)$
  \STATE $\phi_0(x_n) \gets \mathrm{T0Emb}(\tilde{x}_n)$
  \ENDFOR
  \STATE
  \STATE {\color{gray}// build view 1 for unlabeled examples}
  \FOR{$n \mbox{ in } \{1,\ldots, U\}$}
  \STATE {\color{gray}// extract pre-trained DeBERTa representation for $x_n$}
  \STATE $\phi_1(x_n) \gets \mathrm{DeBERTa}(x_n)$
  \ENDFOR
  \STATE
  \STATE {\color{gray}// initialize soft prompt with repeated pad token emb}
  \STATE $p \gets \mathrm{T0Emb}(\texttt{[PAD]})$
  \STATE $h_0(\cdot) \gets \mathrm{T0}((p; p; \ldots; p); \cdot)$
  \STATE
  \STATE {\color{gray}// co-training loop}
  \FOR{$t \mbox{ in } \{0,\ldots, T-1\}$}
      \STATE $\tilde\beta \gets \beta + t\beta'$
      \STATE $L_0^t \gets$ \texttt{GetConfDataCS}($\mathcal{U}, h_{0}, \phi_{0}, \tilde\beta$)
      \STATE $h_{1} \gets$ \texttt{Train}($\phi_{1}, L_0^t$)
      \STATE $L^t_1 \gets$ \texttt{GetConfDataCS}($\mathcal{U}, h_{1}, \phi_{1}, \tilde\beta$)
      \STATE $h_{0} \gets$ \texttt{Train}($\phi_{0}, L_1^t$)
  \ENDFOR
  \STATE {\bfseries return} $(h_0, h_1)$
\end{algorithmic}
\end{algorithm}
\end{minipage}
\end{figure*}

%% file: appendix_deepdive.tex
\begin{figure*}[tb]
    \centering
    \begin{subfigure}{0.48\linewidth}
        \centering
        \includegraphics[width=\linewidth]{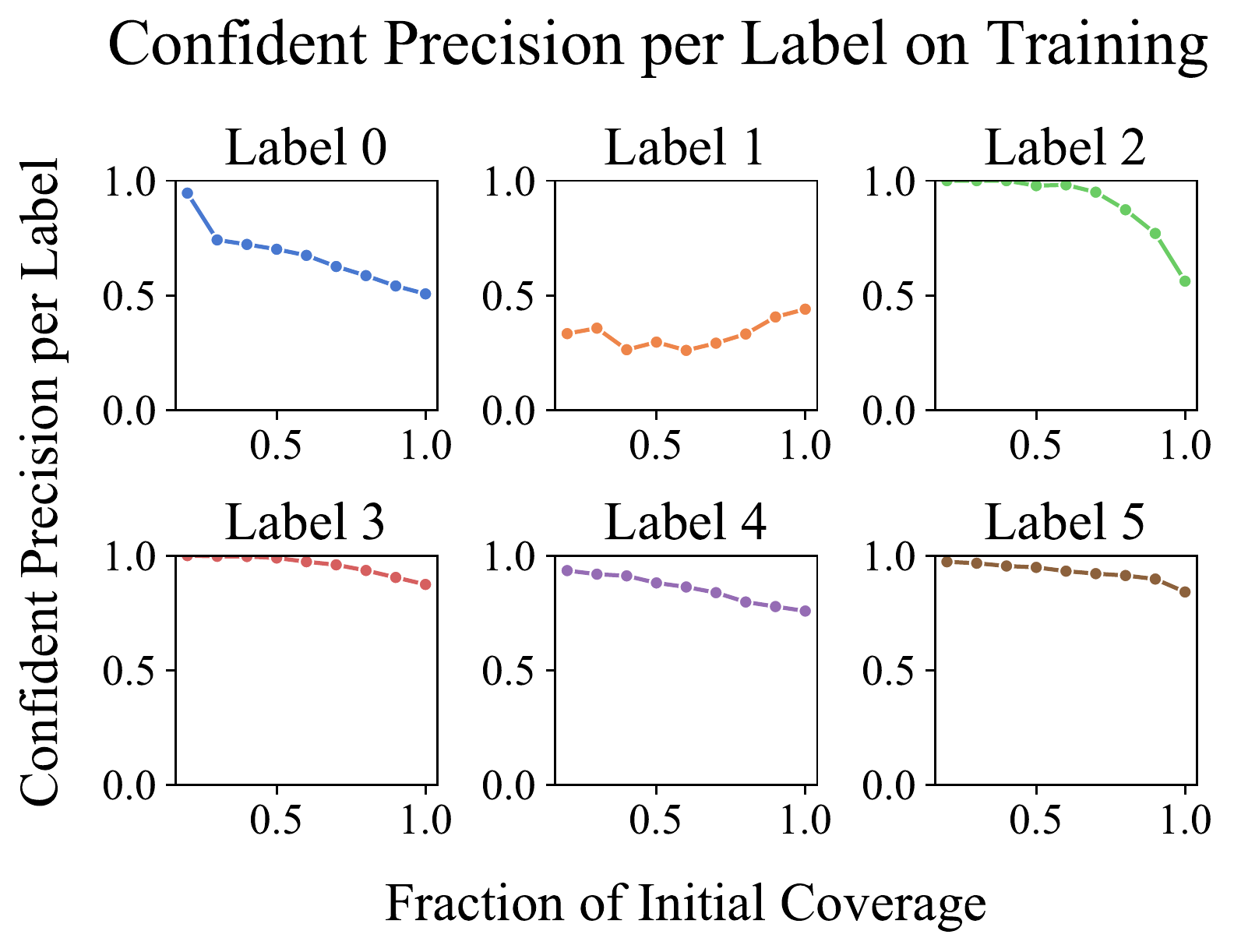}
    \end{subfigure}\hfill%
    \begin{subfigure}{0.48\linewidth}
        \centering
        \includegraphics[width=\linewidth]{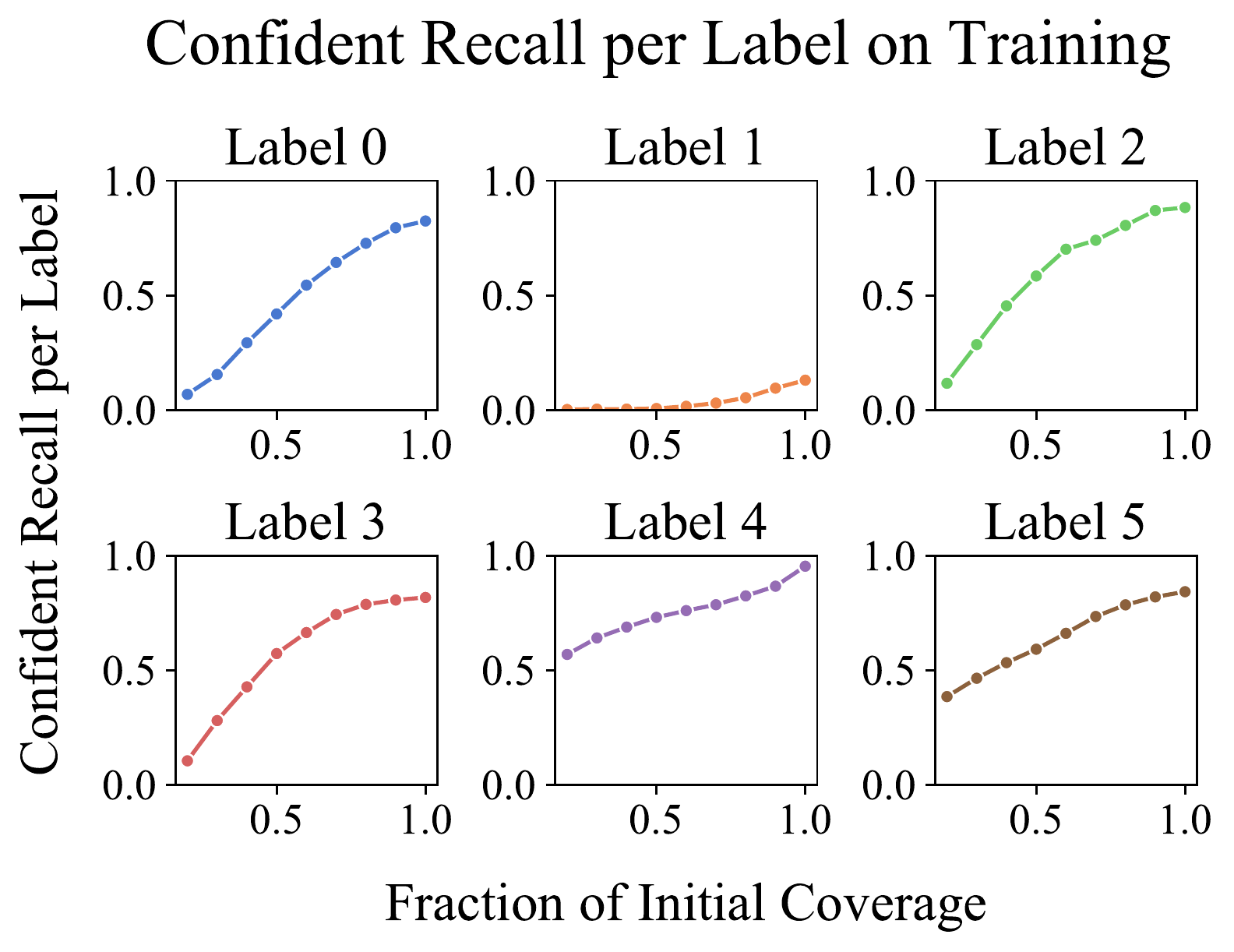}
    \end{subfigure}%
    \vspace{-1em}
    \caption{Precision (left) and Recall (right) versus $\beta$ for each label in the initial confident set $L_0^0$, extracted from the initial label model using the model confidence method (Algorithm \ref{alg:get-conf-data-mc}). The precision of some labels (e.g., 2, 3) begins to decline more sharply after $\beta=0.5$. This gives additional evidence for our choice of $\beta=0.5$: it trades off between the initial precision of $L_0^0$ and the coverage for each label. At smaller values of $\beta$, there are no pseudolabeled examples for label 1 and very few for label 0. At larger values of $\beta$, the precision of the other labels is worse.}
    \label{fig:init_coverage_metrics}
\end{figure*}

\vspace{50mm} %
\section{Additional co-training analysis}
In this section, we provide more information regarding the evolution of $h_0$ and $h_1$ over the co-training iterations for TREC. We focus on the TREC dataset since its 6 classes enable us to investigate more complex co-training dynamics. 

To see the effect of $\beta$ on the quality of the initial confident data $L_0^0$, we plot the precision and recall for each label for different values of $\beta$ in Figure \ref{fig:init_coverage_metrics}. This figure indicates that the tradeoff when choosing $\beta$ is between having high precision for each label (lower $\beta$) and having enough coverage for each label to train on (high $\beta$).

\begin{figure}[tb]
    \centering
    \includegraphics[width=0.9\linewidth]{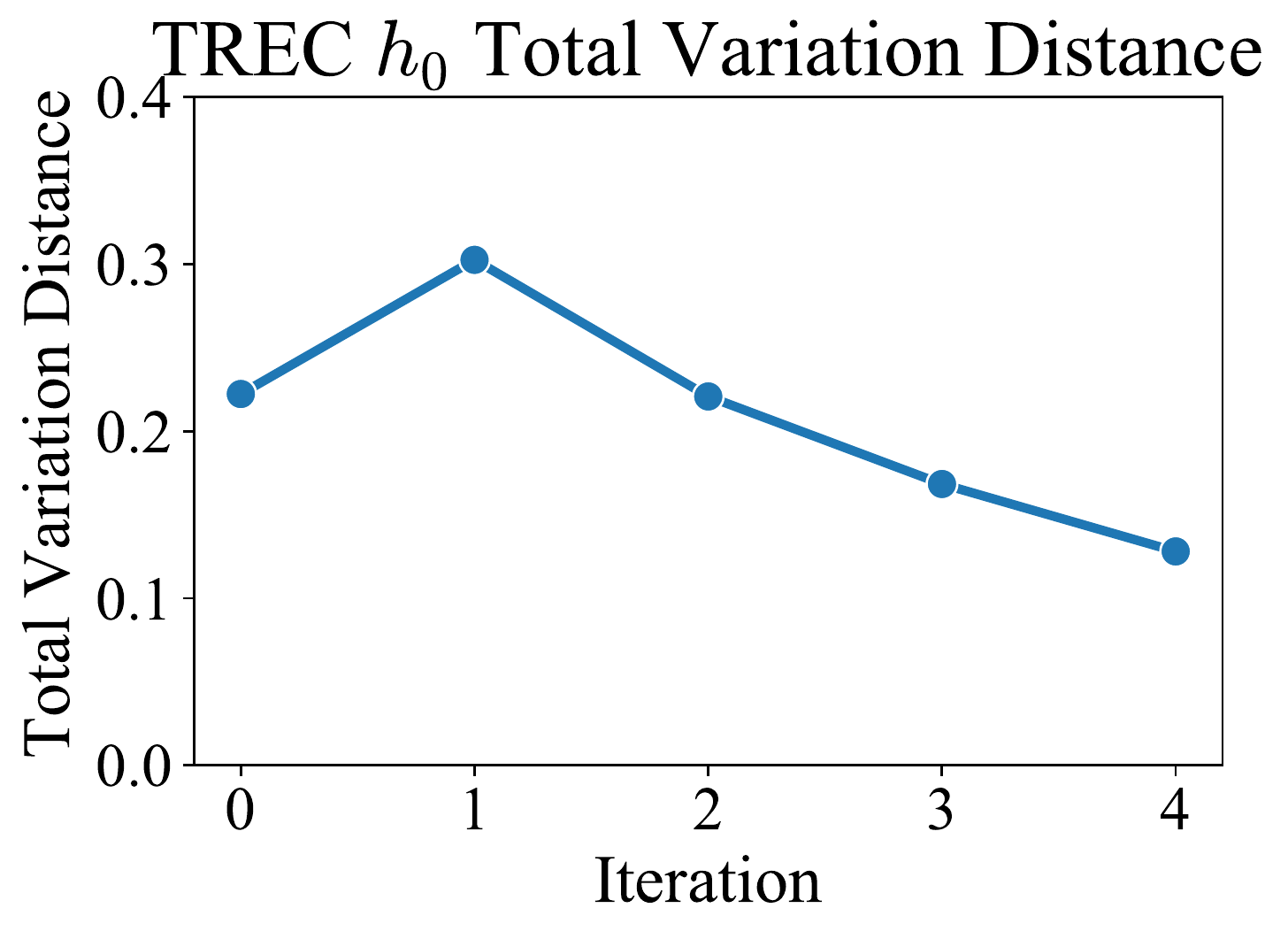}
    \caption{Total variation distance between the true label balance and the label balance estimated from the pseudolabels $L_0^t$ at each iteration. As co-training iterations proceed, the label model automatically learns a balance closer to the true unseen label balance.}
    \label{fig:tvd}
\end{figure}

To show how co-training affects label balance across multiple iterations, we plot the total variation distance between the true label balance and the balance estimated using the pseudolabels in each iteration's confident data $L_0^t$. Figure \ref{fig:tvd} indicates that this distance decreases with co-training iterations, so the label model automatically learns to have a balance closer to the unseen true balance.

\begin{figure*}[tb]
    \centering
    \begin{subfigure}{0.48\linewidth}
        \centering
        \includegraphics[width=\linewidth]{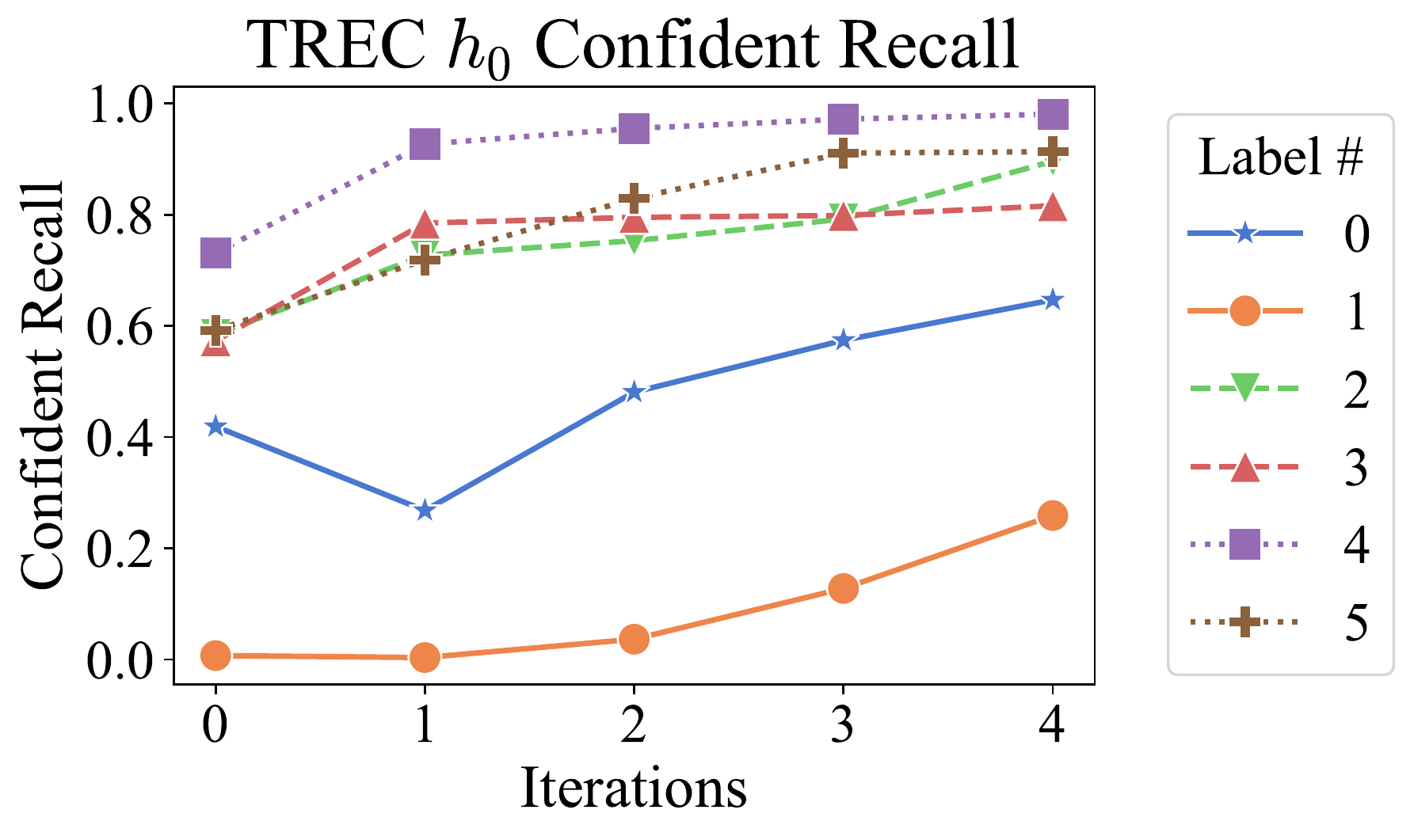}
    \end{subfigure}\hfill%
    \begin{subfigure}{0.48\linewidth}
        \centering
        \includegraphics[width=\linewidth]{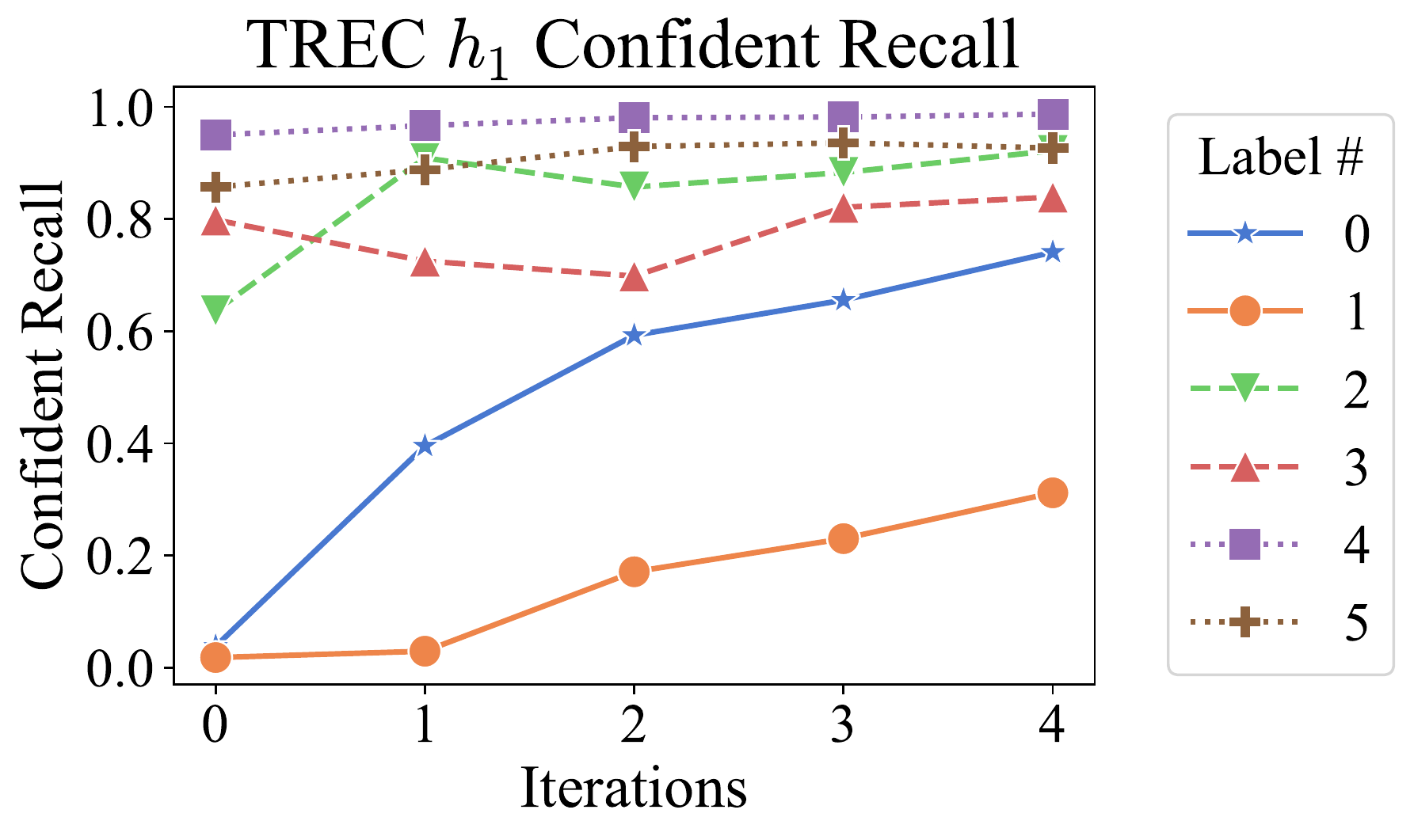}
    \end{subfigure}%
    \vspace{-1em}
    \caption{Recall of the confident pseudolabel set $L_0^t$ (left, extracted from $h_0$ using Algorithm \ref{alg:get-conf-data-mc}) and $L_1^t$ (right, extracted from $h_1$ using Algorithm \ref{alg:get-conf-data-cs}) for each label versus co-training iteration $t$.}
    \label{fig:supp_recall}
\end{figure*}

\begin{figure*}[tb]
    \centering
    \begin{subfigure}{0.48\linewidth}
        \centering
        \includegraphics[width=\linewidth]{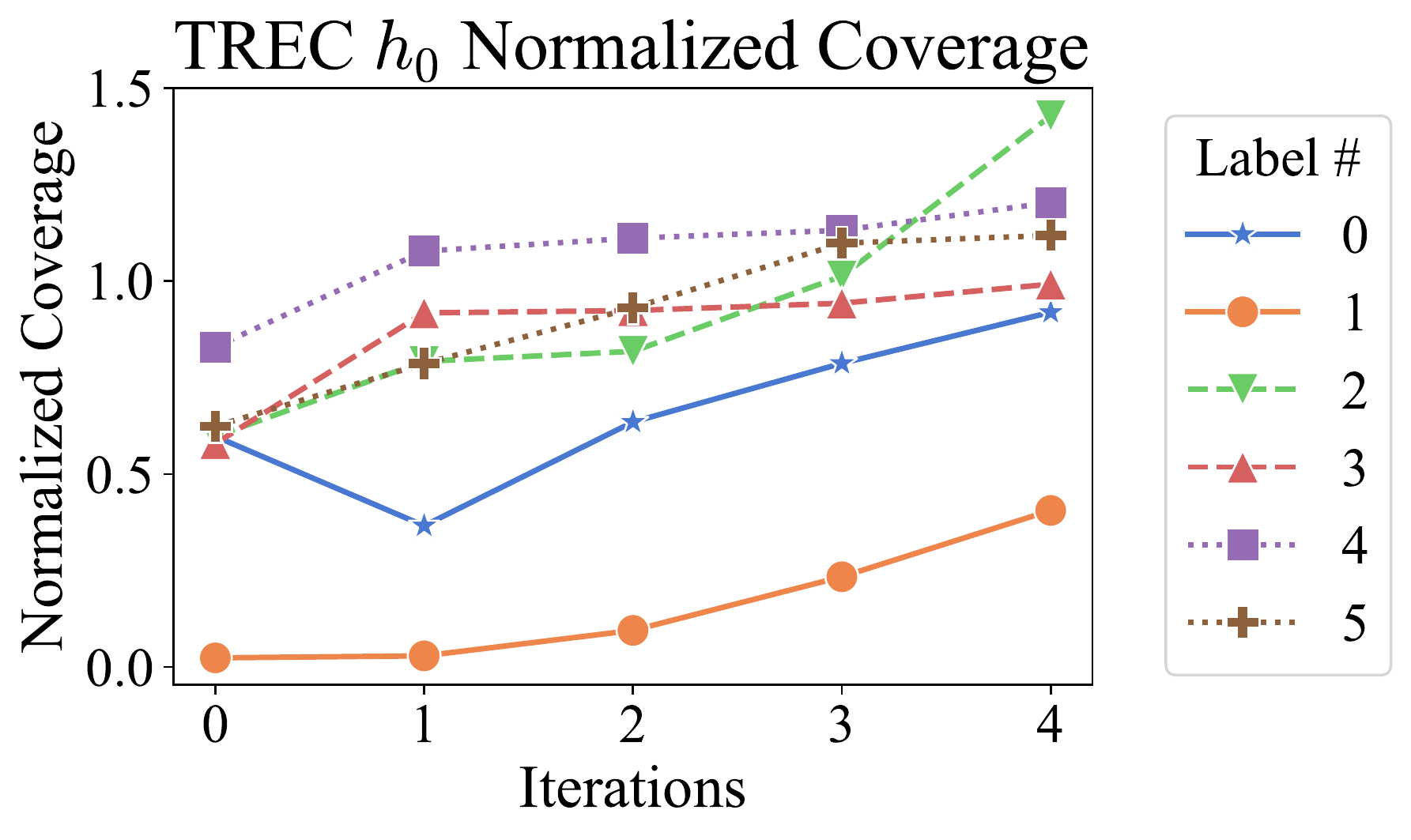}
    \end{subfigure}\hfill%
    \begin{subfigure}{0.48\linewidth}
        \centering
        \includegraphics[width=\linewidth]{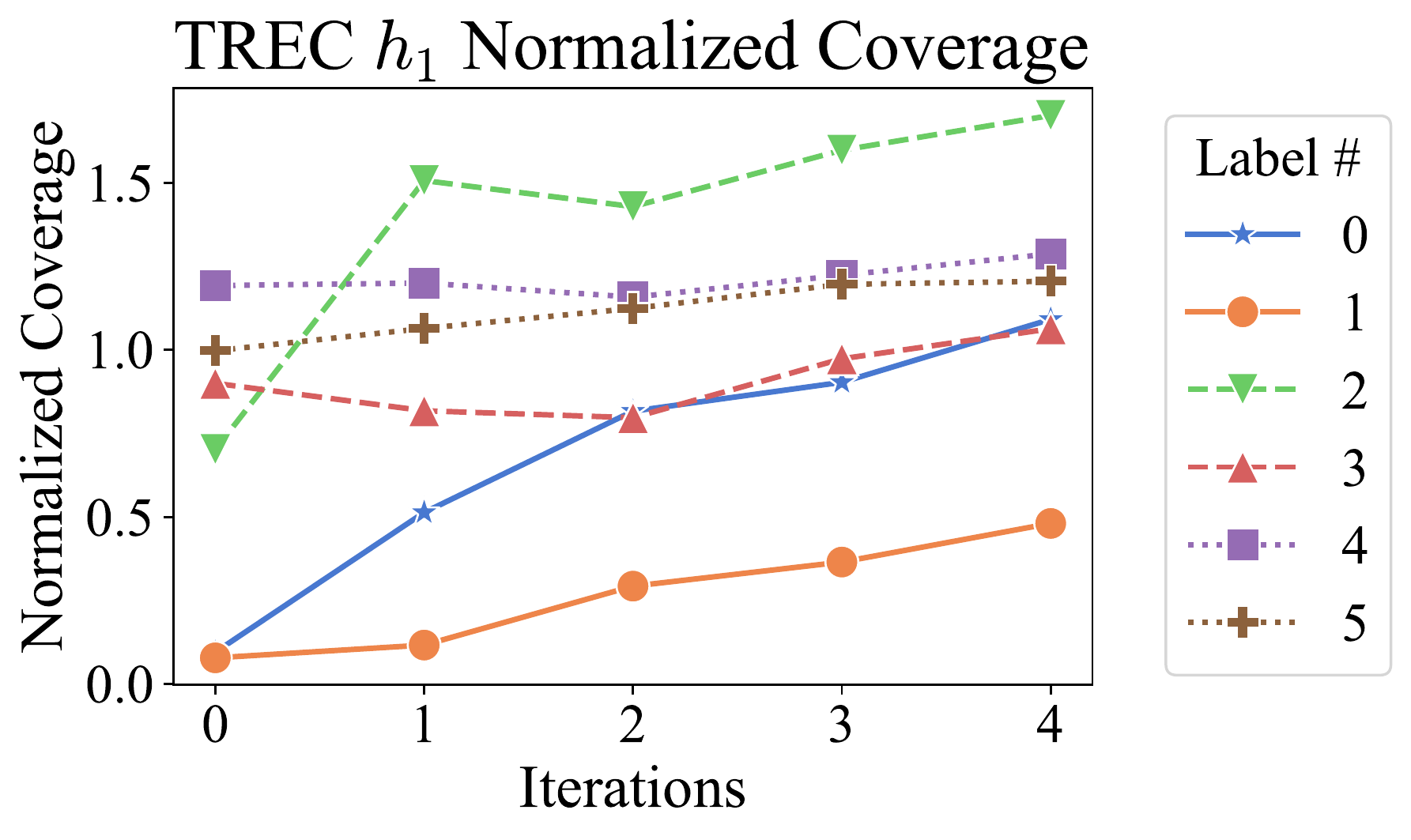}
    \end{subfigure}%
    \vspace{-1em}
    \caption{Normalized coverage of the confident pseudolabel set $L_0^t$ (extracted from $h_0$ using Algorithm \ref{alg:get-conf-data-mc}) for each label versus co-training iteration $t$. Normalized coverage for label $j$ is computed as $|\{(x,\hat{y}) \in L_i^t : \hat{y} = j\}|\ /\ |\{x : y(x) = j\}|$ (the number of examples with confident pseudolabel $j$ divided by the number of examples with true label $j$). This metric decouples the coverage from the precision. The increasing slope of label 1 (left) indicates that $h_0$ adds more confident data for label 1 in the later iterations. Combining this with the label 1 precision versus iteration curve in Figure \ref{fig:supp_precision} (left) indicates that the model adds more confident data for label 1 as it gets more precise, which is the desired behavior. On the other hand, for other labels (e.g. label 4) the rate of confident data addition and the precision stay relatively constant.}
    \label{fig:supp_coverage}
\end{figure*}

\begin{figure*}[tb]
    \centering
    \begin{subfigure}{0.48\linewidth}
        \centering
        \includegraphics[width=\linewidth]{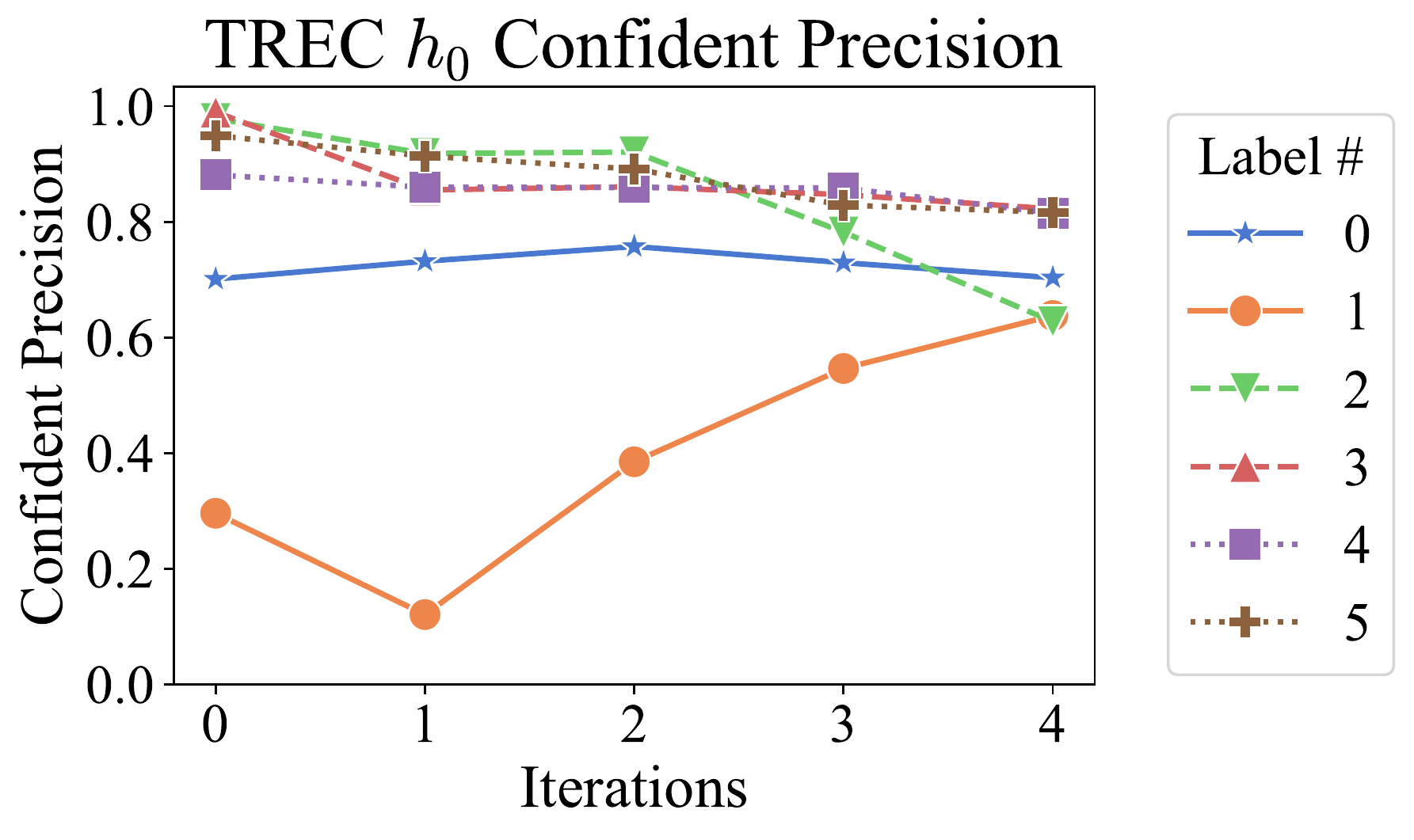}
    \end{subfigure}\hfill%
    \begin{subfigure}{0.48\linewidth}
        \centering
        \includegraphics[width=\linewidth]{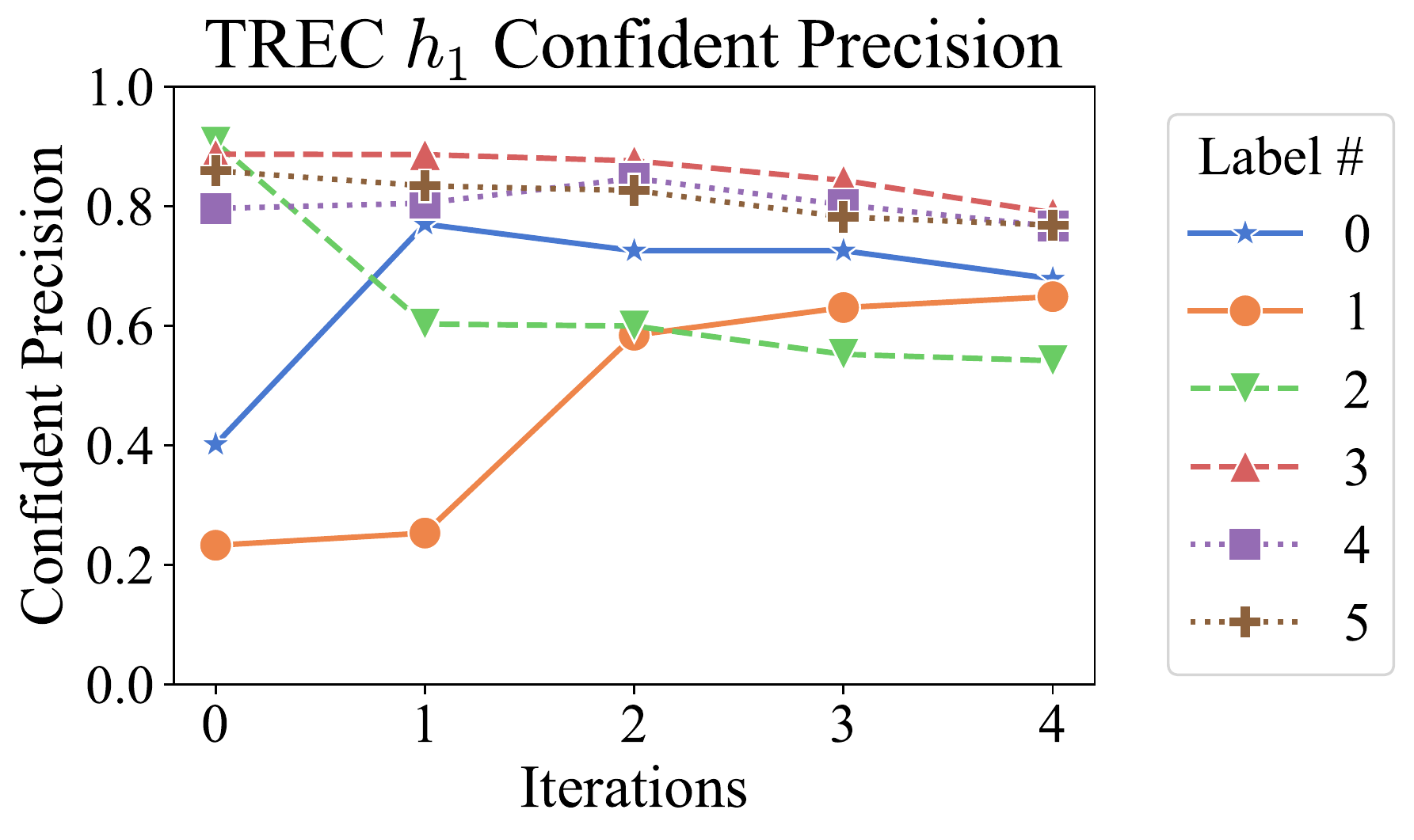}
    \end{subfigure}%
    \vspace{-1em}
    \caption{Precision per label vs co-training iteration, $h_0$ (left---identical to right display of Figure \ref{fig:cotrain-evolution}) and $h_1$ (right). Together with Figure \ref{fig:supp_coverage}, this indicates the two regimes of co-training. For labels 0 and 2-5, the precision decreases or remains the same while the coverage increases roughly linearly. This is the ``coverage-expansion'' regime, where the initial confident pseudolabels are high-precision and the model learns to imperfectly extend that initial signal to the uncovered data with some losses in precision. This regime is present in classical co-training results \citep{balcan2005co}. On the other hand, for label 1, both the coverage \emph{and} the precision increase with the iteration $t$. This is the \emph{pseudolabel-correction} regime, because the models are able to learn to be more accurate than the pseudolabels used to train them (compare $h_0$ precision for label 1 to $h_1$ precision for label 1 in the same iteration---the $h_1$ model is trained on the labels from $h_0$, but is able to select confident data with better precision than those labels).}
    \label{fig:supp_precision}
\end{figure*}

In Figures \ref{fig:supp_recall}, \ref{fig:supp_coverage}, and \ref{fig:supp_precision}, we plot the recall, \emph{normalized coverage}, and precision for each label in $L_0^t$ and $L_1^t$.
The normalized coverage for label $j$ is the number of examples with pseudolabel $j$ divided by the number of examples with true label $j$; it separates coverage from the precision, unlike recall.
By comparing the evolution of label curves in Figure \ref{fig:supp_coverage} and \ref{fig:supp_precision}, we can see that the models tend to add more confident data when they are more precise and add less confident data when they are less precise, which is the desired behavior. Additionally, these figures show two different ways in which co-training works to improve models: ``coverage-expansion'' and ``pseudolabel-correction.''
In the coverage-expansion regime, the precision for a label slightly decreases as iterations increase, but the coverage improves; this regime was predicted by early work on co-training \citep{balcan2005co}. In the pseudolabel-correction regime, both precision and coverage increase, because models are able to learn to be more accurate than the pseudolabels used to train them. \citet{wei2020theoretical} give a theoretical explanation of this in the context of self-training, rather than co-training.

